\def\CLEANVERSION{1}
\documentclass[letterpaper,journal]{IEEEtran}
\usepackage{amsmath,amsfonts,amssymb,amsthm}
\usepackage{algorithmic}
\usepackage{algorithm}
\usepackage{array}
\usepackage[caption=false,font=normalsize,labelfont=sf,textfont=sf]{subfig}
\usepackage{textcomp}
\usepackage{stfloats}
\usepackage{url}
\usepackage{verbatim}
\usepackage{graphicx}
\usepackage{cite}
\usepackage{xcolor}
\usepackage{thmtools}

\newtheorem{lemma}{Lemma}
\newtheorem{assumption}{Assumption}
\newtheorem{definition}{Definition}

\newtheorem{remark}{Remark}
\usepackage{booktabs}
\usepackage{makecell}
\usepackage{multirow}
\usepackage{xspace}
\usepackage{needspace}

\DeclareRobustCommand{\model}{InfMasking\xspace}
\newif\ifannotated
\ifdefined\CLEANVERSION \annotatedfalse \else \annotatedtrue \fi
\ifannotated
  \newcommand{\modify}[1]{\textcolor{red}{#1}}
\else
  \newcommand{\modify}[1]{#1}
\fi

\begin{document}

\title{InfMasking: A Unified Variational Framework for \\Synergistic Multimodal Learning}

\author{Liangjian Wen, Qun Dai, Jiang Duan, Yong Dai, Jianzhuang Liu, Dongkai Wang, Jun Wang, Fuyuan Cao, Zenglin Xu, Zhao Kang
\thanks{
A preliminary version of this work was presented as a spotlight at NeurIPS 2025~\cite{wen2025infmasking}.
}
\thanks{Corresponding authors: Jiang Duan (e-mail: duanj\_t@swufe.edu.cn) and Zhao Kang (e-mail: zkang@uestc.edu.cn).}
\thanks{
L. Wen, Q. Dai, D. Wang, and J. Duan are with the School of Computing and Artificial Intelligence, Southwestern University of Finance and Economics, Chengdu, China. L. Wen is also with the Artificial Intelligence and Digital Finance Key Laboratory of Sichuan Province (e-mail: wlj6816@gmail.com). J. Duan is also with Chengdu Everimaging Science and Technology Co., Ltd. J. Wang is with the School of Management Science and Engineering, Southwestern University of Finance and Economics, Chengdu, China.
}

\thanks{
Y. Dai is with the Beijing Humanoid Robot Innovation Center (X-Humanoid), Beijing, China.
}

\thanks{
J. Liu is with the Shenzhen Institutes of Advanced Technology,
Chinese Academy of Sciences, Shenzhen, China (e-mail: jz.liu@siat.ac.cn).
}
\thanks{
F. Cao is with the School of Computer and Information Technology,
Shanxi University, Taiyuan, China.
}
\thanks{
Z. Xu is with the Artificial Intelligence Innovation and Incubation Institute,
Fudan University and also with the Shanghai Academy of AI for Science,
Shanghai, China.
}
\thanks{
Z. Kang is with the University of Electronic Science and Technology of China,
Chengdu, China.
}
\thanks{
This work was supported by the Major Science and Technology Special Project of the Sichuan Provincial Department of Science and Technology (Grant No. 2024ZDZX0002), the National Natural Science Foundation of China (No. U24A20323), Fundamental Research Funds for the Central Universities (JBK202511081), the Sichuan Provincial Innovation Group Project (Grant No. 2024NSFTD0054), Natural Science Foundation of China under Grant No. 62502397, National Natural Science Foundation of China (NSFC) (72471197), and the Sichuan Provincial Philosophy and Social Science Fund (Grant No. SCJJ25ND091).
}
}

\markboth{IEEE Transactions on Pattern Analysis and Machine Intelligence}%
{Wen \MakeLowercase{\textit{et al.}}: InfMasking for Synergistic Multimodal Learning}


\maketitle

\begin{abstract}
Multimodal models are expected to exploit \emph{synergistic} information, that is, information available only when multiple modalities are observed together. In practice, however, a fusion module can fit the training objective by relying on a dominant modality or on information that modalities share, and the synergy between modalities is often left unused. We propose \model, a unified multimodal learning framework \modify{based} on \emph{Infinite Masking}. Before fusion, \model randomly masks the features of each modality; each sampled mask \modify{produces} a different partial combination of cross-modal evidence \modify{and a} fused representation of \modify{the corresponding} modality \modify{interactions}. All masked fused representations are required to remain consistent with a reference target: the unmasked fused view when training is unsupervised, and the class proxy of the task label when labels are available. Maintaining this consistency across diverse partial combinations discourages reliance on a dominant modality or on shared information alone, and drives the fusion module to learn cross-modal interaction patterns that remain meaningful under partial observations. Since the objective is an expectation over infinitely many masks, we model the masked fused representations as Gaussian and optimize closed-form variational bounds, whose \modify{population-moment} gaps are controlled by the masking-induced covariance \modify{under this model}. On \modify{the evaluated} benchmarks, \model improves \modify{predictive performance} and \modify{yields higher mean PID estimates of task-label synergy in learned modality-specific representations, alongside higher InterSHAP interaction attribution in the trained predictors}. Code is available at \url{https://github.com/brightest66/InfMasking}.
\end{abstract}


\section{Introduction}

\IEEEPARstart{M}{ultimodal} representation learning aims to learn informative joint representations by integrating heterogeneous and complementary information from multiple modalities, such as vision, language, audio, and sensor data~\cite{baltruvsaitis2018multimodal, guo2019deep}.
\modify{A central} challenge \modify{is to design} a fusion mechanism that \modify{integrates} heterogeneous features and \modify{captures} cross-modal interactions.
\modify{Interactions} among modalities can reveal evidence that emerges only from their joint observation and cannot be obtained from any single modality alone.
Attention-based and transformer-based fusion methods have achieved substantial progress in multimodal representation learning by enabling flexible cross-modal interactions~\cite{tsai2019multimodal,xu2023multimodal, nagrani2021attention, wu2024multimodal, dufumier2024align, lu2019vilbert, sun2019videobert, chen2020uniter,li2020oscar}.
However, joint training on multimodal data does not guarantee that all modalities are effectively used\modify{. A} model may exploit redundant correlations or rely on a dominant modality, sometimes performing no better than the strongest unimodal branch while using little of the evidence that emerges only when modalities are combined~\cite{DBLP:conf/colt/SridharanK08, wang2020makes, yao2022modality}.

This difficulty is tied to the information structure of multimodal data~\cite{hu2024survey}. 
Partial Information Decomposition (PID)~\cite{williams2010nonnegative,bertschinger2014quantifying} \modify{offers one information-theoretic way to quantify such joint-only evidence: for specified modality sources and a task target, it separates their} information into redundant, unique, and synergistic components.
Redundant information is comparatively easy to optimize because it can be reinforced through alignment or agreement~\cite{radford2021learning,jia2021scaling}, whereas unique information can be preserved by modality-specific constraints~\cite{guo2025disentangling, wang2024decoupling}.
Synergistic information is more fragile because it emerges from cross-modal interactions and cannot be inferred from any single modality alone.
\modify{Hateful} meme detection \modify{is one example:} harmful semantics \modify{can} arise from \modify{interactions} between visual and textual information rather than from either modality independently~\cite{kiela2020hateful}.
In such cases, the image or the accompanying text may appear innocuous when examined alone, while their combination can encode derogatory, biased, or abusive meanings. 
\modify{Standard multimodal training nevertheless tends to underuse this} synergistic information.

Existing fusion methods partially address this issue. 
In self-supervised settings, 
cross-modal contrastive learning strengthens shared information by aligning views from different modalities, but it often relies on the \emph{multiview redundancy assumption} that shared information across views is sufficient for downstream tasks,
thereby primarily capturing redundant information \cite{DBLP:conf/colt/SridharanK08,DBLP:conf/iclr/Tsai0SM21,tian2020cmc,federici2020learning,dufumier2024align}. 
In supervised settings, 
this problem is further amplified. 
A dominant modality can contribute more to the training objective or converge faster, thereby taking over gradient updates and weakening the optimization of less informative modalities. 
This often leaves the unimodal representations of less informative modalities under-optimized and prevents the model from capturing synergy that emerges only when modalities are combined
~\cite{wang2020makes, peng2022balanced, zhou2023intra,liang2023quantifying}.
Consequently, \modify{a} central \modify{challenge} is to train a fusion mechanism that exploits synergistic interactions \modify{arising from joint observations}, rather than \modify{relying on a dominant} modality \modify{or redundant information alone}.

To address this challenge, we propose \model, a unified multimodal learning framework based on \emph{Infinite Masking}. 
Specifically, \model repeatedly samples masks over \modify{tokens} of the modality-specific representations before fusion.
Each mask \modify{retains} a \modify{different subset of tokens within the modalities}, producing fused representations \modify{from} different \modify{partial cross-modal observations}.
By requiring these masked representations to stay consistent with a learning target \modify{(}an unmasked view in the unsupervised setting or the task label in the supervised setting\modify{),} \model encourages the model to learn interaction patterns that remain meaningful across diverse partial observations.

\model applies to both unsupervised and supervised learning. 
In the unsupervised setting, 
\model aligns masked fused views with their unmasked counterparts to enhance task-agnostic synergistic representations. 
In the supervised setting, 
\model uses a label-proxy objective to require masked fused evidence to remain consistent with the proxy vector of the task label,
thereby promoting decision-level modality collaboration. 
\modify{The} masking operation \modify{remains} the \modify{same in both settings}. \modify{Unsupervised training} preserves view-level information, \modify{while supervised training} preserves \modify{label-relevant decision} evidence. 
Because \modify{exhaustively evaluating} the masking \modify{expectation} is \modify{impractical},
we model the masked fusion distribution as a Gaussian and derive tractable variational bounds: a lower bound for the unsupervised objective and an upper bound for the supervised objective. These bounds serve as surrogate optimization objectives.

Our contributions are threefold:
(1) We formulate \model as a unified multimodal fusion principle that regularizes the fusion module through infinitely many partial modality combinations. The formulation covers both unsupervised representation learning and supervised collaborative optimization.
(2) By modeling masked fusion states as a Gaussian distribution, we derive tractable variational bounds of the infinite masking objective for both the unsupervised and supervised settings. 
Beyond deriving these bounds, we prove \modify{under the Gaussian model} that the slack of each bound is controlled by an anchor-wise (resp.\ class-wise) quadratic form in the masking-induced covariance and vanishes as the masked representations concentrate\modify{;} we \modify{measure the corresponding surrogate gaps by} Monte-Carlo estimation \modify{at} the \modify{tested configurations}. We further show which guarantees survive beyond the Gaussian model: the absolute approximation error is controlled by the masking-induced covariance for arbitrary mask distributions, whereas \modify{a directional sub-Gaussian condition is sufficient for} the one-sided bound property.
(3)
Experiments on multiple multimodal datasets in both unsupervised and supervised settings demonstrate that \model consistently improves performance, enhances robustness to corrupted modalities, and facilitates the discovery of synergistic interactions.
Our code is available at \url{https://github.com/brightest66/InfMasking}.

This paper substantially extends our conference version~\cite{wen2025infmasking} in four aspects.
First, 
it re-examines \model from a unified perspective, formulating a single framework that covers both unsupervised and supervised learning and introducing a label-proxy masked fusion objective.
Second,
it derives a tractable variational upper bound for the supervised objective under a Gaussian model of masked fusion states, extending the bound-based analysis of the conference version beyond the unsupervised case.
Third,
and most importantly from a theoretical standpoint, it establishes, for both the unsupervised and supervised bounds, the tightness \modify{guarantees} described in contribution (2) \modify{under the Gaussian model}, together with Monte-Carlo \modify{evaluation of the corresponding surrogate gaps}, directly \modify{assessing,} rather than merely assuming\modify{,} the \modify{accuracy} of the \modify{Gaussian-based} approximation \modify{at} the \modify{evaluated configurations}.
Fourth,
it provides broader experimental validation of \model in supervised settings.
These extensions show that masked fusion can promote modality collaboration and offer a methodological perspective for future research.

\section{Related Work}

\subsection{Multimodal Interaction and Fusion}
Multimodal fusion aims to integrate heterogeneous observations\modify{,} such as images, text, audio, and sensor data\modify{,} into representations that preserve task-relevant information across modalities, including redundancy shared by multiple modalities, uniqueness specific to individual modalities, and synergy that emerges only from their joint observation~\cite{williams2010nonnegative,bertschinger2014quantifying,liang2023quantifying,liang2024foundations}.
An information-theoretic survey examines multimodal dependency, task-oriented compression, and synergistic interactions through mutual information, information bottleneck, and PID~\cite{wen2026dependency}.
Existing strategies are commonly categorized as 
early fusion, which concatenates modality features before prediction; 
late fusion, which aggregates modality-specific predictions; 
and intermediate fusion, which learns cross-modal interactions through adaptive mechanisms such as gating or attention~\cite{stahlschmidt2022multimodal,baltruvsaitis2018multimodal,arevalo2017gated}.
Recent transformer-based models further strengthen intermediate fusion by enabling token- or feature-level interactions through self-attention and cross-attention~\cite{tsai2019multimodal,xu2023multimodal,barnum2020benefits}.
Cooperation of Experts uses specialized encoders and large-margin cooperation to learn heterogeneous relational patterns~\cite{wang2025cooperation}.
Despite these advances, fusion is still often treated as an architectural operation rather than a training objective: a high-capacity fusion module can fit the training objective by exploiting the most stable or predictive modality, especially when redundant information is available across modalities~\cite{wang2020makes,peng2022balanced, zhang2024multimodal, chaudhuri2025closer}.
In contrast, \model treats fusion as a training objective, exposing the fusion module to a distribution of partial modality combinations so that the learned fusion function must remain meaningful under diverse evidence subsets.

\subsection{Self-Supervised Multimodal Contrastive Learning}
Self-supervised multimodal contrastive learning learns joint representations without labels by exploiting cross-modal correspondences or augmented views.
These methods construct positive pairs from multiple views of the same instance or from naturally paired modalities, and optimize contrastive objectives such as InfoNCE to align semantic representations across modalities~\cite{oord2018representation,chen2020simple,tian2020cmc,radford2021learning,jia2021scaling}.
From a multiview perspective, their effectiveness is often explained by the assumption that different views preserve shared task-relevant information~\cite{DBLP:conf/colt/SridharanK08,DBLP:conf/iclr/Tsai0SM21,federici2020learning,wen2024mveb}.
Beyond contrastive estimators, mutual information gradient estimation has also been explored for representation learning~\cite{DBLP:conf/iclr/WenZHZX20}.
Multiview clustering methods impose information bottleneck constraints at granular and sample levels~\cite{lou2026granular} or combine self-supervised view weighting with clustering-structure consistency~\cite{lou2026structure}. Other methods combine pseudo-label-guided contrastive learning with feature- and cluster-level correlation learning~\cite{hu2025pseudolabel}, or jointly model uncertainty in the data, model, and predictions through full-stage trusted multimodal clustering~\cite{hu2026trust}.
Another line of multimodal self-supervised learning adopts masked modeling or autoencoding objectives, where models reconstruct missing content from partially observed inputs and thereby learn robust cross-modal representations~\cite{he2022masked,bachmann2022multimae,lin2022dual}.
However, both view agreement and reconstruction can still be dominated by redundant or easily recoverable correlations, and neither objective necessarily requires the fusion module to expose interaction patterns that only appear after modalities are combined.
Several recent methods explicitly move beyond pure redundancy~\cite{hu2024multiview,zhang2023multi,liang2023factorized,dufumier2024align}.
Factorized contrastive learning decomposes shared and unique factors to better model multi-view information~\cite{liang2023factorized}, 
while contrastive multimodal interaction learning introduces information-theoretic objectives to capture redundant, unique, and synergistic components~\cite{dufumier2024align}. 
HRIL models higher-order interactions through cross-moment tensors, Tucker decomposition, and synergy-aware regularization~\cite{dai2026hril}.
The conference version of \model further showed that masked and unmasked fused views can be aligned to enhance task-agnostic synergistic representations~\cite{wen2025infmasking}.
\modify{\model generalizes} this \modify{masked-fusion approach to both learning settings}, \modify{constraining} the same distribution of masked fusion states \modify{through} view-level consistency in the unsupervised setting or label-proxy decision evidence in the supervised setting.

\subsection{Supervised Multimodal Learning}
Supervised multimodal learning introduces task labels to the fusion problem, but empirical risk minimization alone does not ensure effective modality collaboration~\cite{wang2020makes,peng2022balanced}.
Task-specific approaches include semantic-anchor-guided data augmentation for low-resource multimodal information extraction~\cite{guo2026sama} and knowledge-induced multimodal reasoning for web agents~\cite{guo2026webcogreasoner}.
The emergence of modality competition, where one modality rapidly minimizes the training loss, prevents the model from fully exploiting cross-modal interactions~\cite{huang2022modality,yao2022modality}.
This failure mode is particularly harmful when task-relevant information is distributed across modalities, because a dominant-modality shortcut can suppress the interactions that make multimodal learning effective.
Existing methods address this issue from different perspectives.
Multiloss~\cite{vielzeuf2018centralnet} introduces auxiliary losses on modality-specific branches to preserve unimodal discriminability, 
while MMCosine~\cite{xu2023mmcosine} equalizes modality influence by standardizing features and classifier weights.
Other approaches balance multimodal optimization by modulating gradient contributions, coordinating modality-specific objectives, or incorporating alignment regularization~\cite{yao2022modality,peng2022balanced,wei2024mmpareto,wei2024diagnosing}.
Although these strategies improve robustness and preserve unimodal discriminability, they balance modality contributions without explicitly requiring partial multimodal evidence to support the task decision, and thus may be insufficient for tasks that depend on cross-modal interactions.
MSCR addresses both modality imbalance and synergy learning by combining anti-dominance and synergy objectives with sample-adaptive gating~\cite{chen2026mscr}.
\modify{To} explicitly \modify{promote} label-consistent cross-modal synergy\modify{,} \model \modify{requires} each partially masked fused representation to remain consistent with the label proxy\modify{. This objective} encourages complementary interactions among modalities while preserving the predictive strengths of individual modalities.

\section{\modify{Preliminaries}}
\label{sec:prelim}

\textbf{Problem Setup.}
Let \(X=(X_1,\ldots,X_n)\) denote an \(n\)-modal input, where \(X_m\) is the observation from the \(m\)-th modality, such as text, image, or audio.
Let \(Y\) denote the downstream task label. For each modality, a modality-specific encoder \(f_m(\cdot;\theta_m)\) maps \(X_m\) to a latent feature \(h_m=f_m(X_m;\theta_m)\), where \(\theta_m\) denotes the encoder parameters. 
A fusion module \(g_{\phi}\) then takes the set of modality features \(H=\{h_m\}_{m=1}^{n}\) as input and produces a joint representation \(Z=g_{\phi}(H)\), where \(\phi\) denotes the fusion parameters. 
The goal is to learn \modify{the representation parameters} \(\theta=\{\modify{\theta_1,\ldots,\theta_n,\phi\}}\) such that \(Z\) preserves the task-relevant information contained in the multimodal input and supports downstream prediction. 
Ideally, \(Z\) should be informative for \(Y\), \textit{i.e.}, \(I(Z;Y)= I(X;Y)\).
\modify{For brevity we also write \(f_\theta=g_\phi\circ\{f_m\}_{m=1}^{n}\) for the composition of the modality-specific encoders with the fusion module, so that \(Z_\theta=f_\theta(X)=g_\phi(\{f_m(X_m;\theta_m)\}_{m=1}^{n})\) is the fused representation of \(X\) and, for an augmented view \(X'\), \(Z'_\theta=f_\theta(X')=g_\phi(\{f_m(X'_m;\theta_m)\}_{m=1}^{n})\) is the fused representation of that view, computed by the same encoders and fusion module. Here \(\theta\) collects the encoder and fusion parameters of this representation map; any additional classifier or projection-head parameters are specified separately. The prime denotes an augmented input view, not a different parameter set, and the parameter subscript is omitted when it is clear from context.}

\subsection{Multimodal Interaction}
Multimodal fusion \modify{must model} interactions among \modify{modalities as well as aggregate their features}.
For two modalities \(X_1\) and \(X_2\),
Partial Information Decomposition (PID)~\cite{williams2010nonnegative,bertschinger2014quantifying} provides a framework to characterize multimodal interactions by decomposing the information that multiple modalities provide about a target variable \(Y\) into redundant, unique, and synergistic terms:
\begin{equation}
    I(X_1,X_2;Y)=R+S+U_1+U_2,
    \label{eq:pid}
\end{equation}
where \(R\) denotes redundancy shared by both modalities, \(U_i\) denotes information unique to \(X_i\), and \(S\) is available only through their joint observation.
This decomposition is governed by consistency equations derived from the chain rule of mutual information:
\begin{equation}
    I(X_i;Y)=R+U_i, \quad
    I(X_1;X_2;Y)=R-S.
    \label{eq:pid_consistency}
\end{equation}
\modify{The} synergistic information \(S\) is \modify{particularly relevant to} multimodal learning \modify{because} it captures patterns that arise only from \modify{combined} modalities and \modify{cannot be inferred} from any single modality in isolation.

\subsection{Contrastive Multimodal Learning}

This section builds upon Section~2 of~\cite{dufumier2024align}, on which \model is developed, to keep notation consistent.
In the unsupervised formulation of multimodal contrastive learning, 
task labels \(Y\) are unavailable during pre-training. 
To make the objective tractable without label supervision, the following assumption is introduced in~\cite{dufumier2024align}:
\begin{assumption}[\textit{Minimal label-preserving multimodal augmentations}]
        \label{assumption: label-preserving}
        We assume the existence of \(\mathcal{T}^*\), a set of multimodal augmentations such that for any \(t \in \mathcal{T}^*\) and \(X' = t(X)\), we have \(I(X;X') = I(X;Y)\).
\end{assumption}
Assumption~\ref{assumption: label-preserving} admits a broader class of augmentations than commonly used unimodal transformations, which perturb individual modalities independently.
This broader formulation permits joint multimodal augmentations that act on cross-modal settings while preserving label-relevant information.
For an augmented view \(X'=t(X)\), where \(t \in \mathcal{T}^*\), we denote its representation by \(Z'_\theta = f_\theta(X')\).
Since \(Z_{\theta}'\) is obtained from \(X'\), the variables satisfy the Markov chain \(X\rightarrow X'\rightarrow Z'\); together with \(Z'\rightarrow X\rightarrow Z\), the data processing inequality yields
\begin{equation}
    I(Z;Z')\le I(X;Z')\le I(X;X').
    \label{eq:dpi_chain}
\end{equation}
\modify{Following CoMM}~\cite{dufumier2024align}\modify{, maximizing the mutual information between paired representations encourages preserving information shared by the augmented views.}

\begin{figure*}[!ht]
    \centering
    \includegraphics[width=0.95\textwidth,height=0.39\textheight,keepaspectratio]{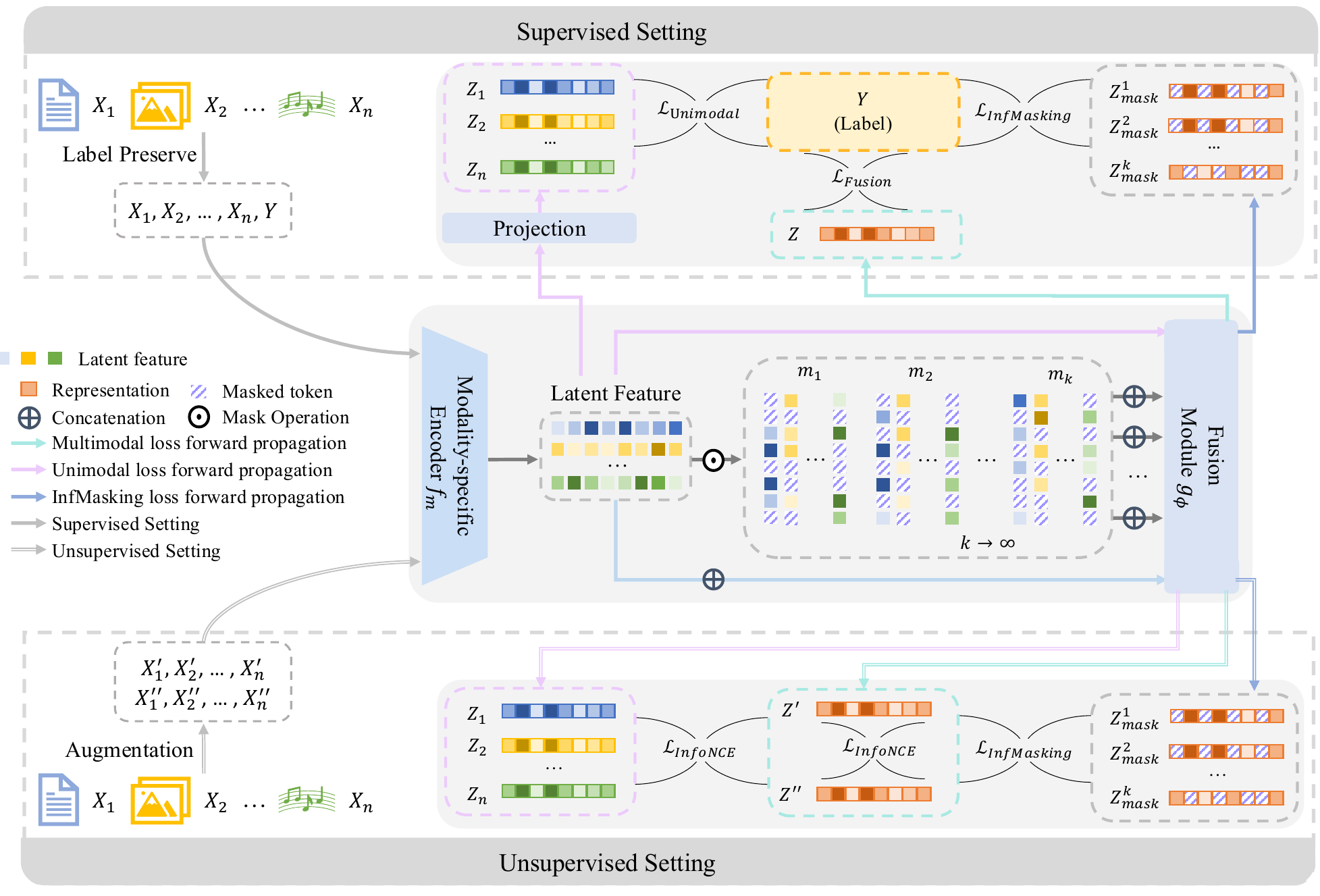}
    \caption{Overview of the proposed \model framework. 
    Given \(n\) multimodal inputs \(X=\{X_1, X_2, \ldots, X_n\}\), \model \modify{uses} label-proxy supervision in the supervised setting (upper) and \modify{augmented-view alignment} in the unsupervised setting (lower).
    Both settings share the same backbone architecture.
    Modality-specific encoders \(\modify{f_m}\)\modify{, \(m=1,\ldots,n\),} extract latent features, and the infinite masking operator samples subsets \modify{of their tokens} before fusion. 
    The fusion module \(g_\phi\) then produces the complete multimodal representation \(Z\) from \modify{unmasked modality features} and masked fused representations \(Z_{\mathrm{mask}}^1, Z_{\mathrm{mask}}^2, \ldots, Z_{\mathrm{mask}}^k\) from sampled \modify{combinations of retained tokens across modalities}.
    In the unsupervised setting, the fusion module \(g_\phi\) additionally produces unimodal representations \(Z_1, Z_2, \ldots, Z_n\) from individual modalities, whereas in the supervised setting the corresponding unimodal branches are obtained by projecting modality-specific latent features.
    }
    \label{fig:flow_chart}
\end{figure*}

Based on the above, \modify{we recall Lemmas 2 and 3 of CoMM}~\cite{dufumier2024align}\modify{, which} summarize the role of contrastive alignment in modeling multimodal interactions.

\begin{lemma}
    By optimizing \(f_\theta\) to maximize \(I(Z_\theta;Z_\theta^{\prime})\), and assuming an expressive enough network \(f_\theta\), we have at optimum:
    \begin{equation}
        I(Z_{\theta^{\star}};Z'_{\theta^{\star}})=I(X;X').
    \end{equation}
    \label{lem:common_rep}
\end{lemma}
\modify{Following CoMM}~\cite{dufumier2024align}\modify{, we use paired-view alignment to motivate learning multimodal representations that reflect cross-modal interactions.}

\modify{For the single-modality input used below, the notation \(f_\theta(X_i)\) denotes \(g_\phi(\{f_i(X_i;\theta_i)\})\): the corresponding encoder supplies its modality features to the same fusion module. This is the single-modality restriction of the representation map in the unsupervised formulation, rather than a supervised projection-head output.}

\begin{lemma}
        Let \(f_{\theta^{\star}}\) be optimal, i.e., \(f_{\theta^{\star}}\) maximizes \(I(Z_\theta;Z'_\theta)\). Then, we have the equality \(I(Z'_{\theta^{\star}};Y)=I(X';Y)\). If we consider the special case \(\mathcal{T}=\{t_i\}\) such that \(X'=t_i(X)=X_i\) and \(Z'_{\theta^{\star}}=f_{\theta^{\star}}(X_i)=Z_i\) for \(i\in\{1,2\}\), then it follows:
        \begin{equation}
            I(Z_i;Y)=I(X_i;Y)=R+U_i.
        \end{equation}
        \label{lem:unimodal_rep}
\end{lemma}
Lemma~\ref{lem:unimodal_rep} provides the theoretical motivation for introducing unimodal contrastive branches that preserve the label-relevant information carried by an individual modality.
Maximizing \(I(Z_i;Z'_{\theta^{\star}})\) thus lets the unimodal representation capture both \(R\) and \(U_i\)\modify{. \model addresses} the remaining difficulty\modify{:} strengthening interaction patterns that require modalities to be fused.

\section{Method}

\subsection{\modify{Overview}}
\label{sec:overview}
\modify{We propose \model, a unified multimodal learning framework based on the principle of} \emph{\modify{Infinite Masking}}\modify{,} 
\modify{which can be applied to both unsupervised and supervised learning to promote synergistic interactions and modality collaboration.}
\modify{The architecture of \model is illustrated in Fig.}~\modify{\ref{fig:flow_chart}} 
\modify{and consists of two stages: modality-specific encoding and multimodal fusion.} 
\modify{For each modality \(X_m\), a modality-specific encoder \(f_m\) extracts a latent feature \(h_m\).}
\modify{We then sample \(k\) binary masks and apply them to the set of modality features \(H=\{h_m\}_{m=1}^{n}\) before fusion.}
\modify{Concretely, the masking operator applies a binary mask \(\boldsymbol{m}_m\) to the token sequence \(h_m\) of each modality \(m\), zeroing each token independently with probability \(\rho\) (the} \emph{\modify{masking ratio}}\modify{) to form \(\boldsymbol{m}_m\odot h_m\). The mask is applied to tokens after modality-specific encoding and before fusion, rather than by selecting whole modalities for removal.}
\modify{Collecting the masked modality features as \(\widetilde{H}=\{\boldsymbol{m}_m\odot h_m\}_{m=1}^{n}\), the fusion module maps them to a masked fused representation}
\begin{equation}
    \modify{Z_{\mathrm{mask}} = g_{\phi}(\widetilde{H}),}
    \label{eq:masking_op}
\end{equation}
\modify{so that each sampled set of per-modality masks exposes the same fusion module \(g_{\phi}\) to a different ``partial modality combination,'' consisting of the tokens retained within each modality. Whole-modality dropout is evaluated separately as a test-time corruption in the robustness evaluation.}
\modify{Sampling \(k\) masks yields \(k\) masked fused views \(Z_{\mathrm{mask}}^1,\ldots,Z_{\mathrm{mask}}^k\) (throughout, uppercase \(Z\) denotes a representation as a random variable and lowercase \(z\) its realization); letting \(k\to\infty\) defines a distribution over masked fusion states.}
\modify{We define the resulting training objective as follows.}
\begin{definition}[Infinite Masking]
\label{def:infmasking}
\modify{Let \(q_{\rho}\) denote the distribution of the per-modality masks \(\{\boldsymbol{m}_m\}_{m=1}^{n}\), under which each token is zeroed independently with probability \(\rho\), and let \(Z_{\mathrm{mask}}=g_{\phi}(\widetilde{H})\) be the resulting masked fused representation.}
\emph{\modify{Infinite Masking}} \modify{trains the encoders and the fusion module on the expectation of a consistency objective over this mask distribution,}
\begin{equation}
    \mathcal{\modify{L}}_{\mathrm{InfMasking}}
    \modify{=}\mathbb{\modify{E}}_{\mathrm{mask}\sim q_{\rho}}\big[\delta(Z_{\mathrm{mask}},T)\big]\modify{,}
    \label{eq:infmasking_def}
\end{equation}
\modify{where the reference target \(T\) and the discrepancy \(\delta\) are instantiated as the unmasked fused view with the negative InfoNCE alignment in the unsupervised setting (applied to each augmented view; Eq.}~\modify{(\ref{eq:mask_loss})), and as the class-proxy set with the cross-entropy loss of the target class in the supervised setting (Eq.}~\modify{(\ref{eq:sup_infmasking})).}
\end{definition}
\modify{The same fusion module \(g_\phi\) produces the complete representation \(Z\) from unmasked features and the masked representations \(Z_{\mathrm{mask}}^1, Z_{\mathrm{mask}}^2, \ldots, Z_{\mathrm{mask}}^k\) from the sampled token combinations.}
\modify{In the unsupervised setting, the fusion module additionally produces unimodal representations \(Z_1, Z_2, \ldots, Z_n\) from individual modalities, whereas in the supervised setting the corresponding unimodal branches are obtained by projecting modality-specific latent features.}
\modify{In unsupervised learning, where labels are unavailable, augmented views provide the alignment reference; in supervised learning, the class proxies of the task label provide the training target.}
\modify{The training procedures of the two regimes are summarized as pseudocode in the supplementary material.}

\subsection{Infinite Masking: Unsupervised View Alignment}
For unlabeled data, we first apply data augmentation to the modality inputs \(X=\{X_1, X_2, \ldots, X_n\}\) to obtain views \modify{for contrastive alignment}.
Given an augmented fused view \(Z'\), we use the InfoNCE~\cite{oord2018representation} estimator to quantify the mutual information between \(Z\) and \(Z'\):

\begin{equation}
    \resizebox{0.99\linewidth}{!}{$\displaystyle
    \widehat{I}_{\mathrm{NCE}}(Z,Z')
    =
    \mathbb{E}\left[
    \log
    \frac{\exp(z^\top z_{\mathrm{pos}}'/\tau)}
    {\exp(z^\top z_{\mathrm{pos}}'/\tau)+\sum_{z_{\mathrm{neg}}'}\exp(z^\top z_{\mathrm{neg}}'/\tau)}
    \right].
    $}
    \label{eq:infonce_estimator}
\end{equation}

Based on Lemma~\ref{lem:common_rep} and Lemma~\ref{lem:unimodal_rep}, CoMM~\cite{dufumier2024align} introduces a contrastive multimodal interaction objective to align fused views while preserving modality-specific information through unimodal alignment. Its training objective is
\begin{equation}
    \begin{aligned}
    \mathcal{L}_{\mathrm{CoMM}}
    =
    &-\widehat{I}_{\mathrm{NCE}}(Z',Z'')\\
    &-\sum_{i=1}^{n}\frac{1}{2}
    \left[
    \widehat{I}_{\mathrm{NCE}}(Z_i,Z')
    +
    \widehat{I}_{\mathrm{NCE}}(Z_i,Z'')
    \right].
    \end{aligned}
    \label{eq:comm_loss}
\end{equation}
\modify{Under the ideal optima and the conditions of Lemmas}~\modify{\ref{lem:common_rep} and}~\modify{\ref{lem:unimodal_rep}, fused-view alignment} is \modify{motivated by preserving task-relevant joint information}, \modify{while} the unimodal \modify{alignment} terms \modify{motivate retaining the label information available to each modality}. \modify{The contrastive objective does not directly estimate or decompose the PID components} of the \modify{learned representations}.

To strengthen \modify{the} synergistic interactions essential for multimodal learning, \model applies infinite masking before fusion\modify{, producing} masked fused views \modify{from} diverse partial modality combinations\modify{. Aligning} each \modify{masked view} with its unmasked counterpart encourages the fusion module to learn interaction patterns that remain meaningful across these combinations.
\modify{This} objective \modify{regularizes the fusion module across partial modality combinations without} directly \modify{optimizing} an estimator of the synergistic component \(S\) in Eq.~(\ref{eq:pid})\modify{. We assess} its effect through the controlled Trifeature benchmark (Section~\ref{sec:unsup_exp})\modify{, partial-information-decomposition estimates on the learned representations (Section}~\modify{\ref{sec:synergy_quant}),} and InterSHAP analysis (Section~\ref{sec:additional_exp}).
To capture a broad range of synergistic interaction patterns, we repeat the masking process \(k\) times and define the masking loss \(\mathcal{L}_{\mathrm{InfMasking}}^{\mathrm{unsup}}\) as
\begin{align}
     -\lim _{k\rightarrow \infty}\frac{1}{k}\sum_{i=1}^k\left[\widehat{I}_{\text{NCE}}(\modify{Z^{\prime}, Z^{\prime~i}_{\mathrm{mask}}}) +\widehat{I}_{\text{NCE}}(\modify{Z^{\prime\prime}, Z^{\prime\prime~i}_{\mathrm{mask}}})\right] \nonumber
     \\=  -\mathbb{E}_{\mathrm{mask}}\left[ \widehat{I}_{\text{NCE}}(\modify{Z^{\prime}, Z^{\prime}_{\mathrm{mask}}}) + \widehat{I}_{\text{NCE}}(\modify{Z^{\prime\prime}, Z^{\prime\prime}_{\mathrm{mask}}}) \right].
    \label{eq:mask_loss}
\end{align}
This loss instantiates Definition~\ref{def:infmasking} with the unmasked fused view as the reference target.
Increasing \(k\) exposes the model to a broader range of synergistic interactions, but the expectation in Eq.~(\ref{eq:mask_loss}) cannot be evaluated exactly, and a finite-sample Monte-Carlo average provides only a stochastic estimate of it.
To obtain a tractable approximation, inspired by~\cite{cai2020joint}, we assume that, over the random masks, the masked fused representation \(z^{\prime}_{\mathrm{mask}}\) follows a Gaussian distribution:
\(z^{\prime}_{\mathrm{mask}} \sim \mathcal{N}(\boldsymbol{\mu}', \boldsymbol{\Sigma}')\),
where \(\boldsymbol{\mu}'\) and \(\boldsymbol{\Sigma}'\) denote the mean vector and covariance matrix, estimated in practice as the empirical mean and covariance of the \(k\) sampled masked representations.
This assumption provides a compact statistical characterization of the masked fused representations:
the mean vector summarizes their shared semantic content, whereas the covariance matrix characterizes the variation induced by different partial modality combinations.
Under this assumption, we derive the following variational lower bound for the infinite masking objective. 
The complete proof is provided in the supplementary material.
\begin{restatable}{lemma}{unsupInfMasking}
    Let $\boldsymbol{\mu}'$ and $\boldsymbol{\Sigma}'$ denote the mean vector and covariance matrix of the Gaussian approximation to $z^{\prime}_{\mathrm{mask}}$, respectively, and define $\mathcal{J}'=\mathbb{E}_{\mathrm{mask}}[\widehat{I}_{\text{NCE}}(\modify{Z^{\prime}, Z^{\prime}_{\mathrm{mask}}})]$. Then
    \begin{alignat}{2}
        \mathcal{J}'
        &\geq
        \mathbb{E}_{z'\sim p(Z')}\Bigg[
        \frac{z'^{\top}\boldsymbol{\mu}'}{\tau}
        &&{}-
        \log\Bigg(
        \exp\left(
        \frac{z'^{\top}\boldsymbol{\mu}'}{\tau}
        +
        \frac{z'^{\top}\boldsymbol{\Sigma}'z'}{2\tau^2}
        \right)
        \nonumber\\
        &&&{}+
        \sum_{z'_{\text{neg}}}
        \exp(z'^{\top}z'_{\text{neg}}/\tau)
        \Bigg)
        \Bigg].
    \end{alignat}
    \label{lemma:unsup_inf}
\end{restatable}
This lower bound provides \model with a tractable approximation to the mutual information between masked and unmasked fused representations without exhaustively sampling all possible masks.
\modify{Under the Gaussian model, its} slack is controlled by the masking-induced covariance and vanishes as the masked representations concentrate; we formalize this property, together with its supervised counterpart, in Section~\ref{sec:tightness} and \modify{measure the observed surrogate gaps} in Section~\modify{\ref{sec:tightness_exp}}.
The full unsupervised objective of \model combines the contrastive interaction baseline with the masking term to further promote synergistic interactions:
\begin{equation}
    \begin{aligned}
    \mathcal{L}_{\mathrm{unsup}}
    =
    &-\widehat{I}_{\mathrm{NCE}}(Z',Z'')\\
    &-\sum_{i=1}^{n}\frac{1}{2}
    \left[
    \widehat{I}_{\mathrm{NCE}}(Z_i,Z')
    +
    \widehat{I}_{\mathrm{NCE}}(Z_i,Z'')
    \right] \\
    &-\mathbb{E}_{\mathrm{mask}}\left[ \widehat{I}_{\text{NCE}}(\modify{Z^{\prime}, Z^{\prime}_{\mathrm{mask}}}) + \widehat{I}_{\text{NCE}}(\modify{Z^{\prime\prime}, Z^{\prime\prime}_{\mathrm{mask}}}) \right].
    \end{aligned}
    \label{eq:full_unsup_loss}
\end{equation}
In practice, the masked term is optimized through the variational lower bound in Lemma~\ref{lemma:unsup_inf}, with \(\boldsymbol{\mu}'\) and \(\boldsymbol{\Sigma}'\) estimated from a finite set of \(k\) sampled masks.

\subsection{Infinite Masking: Supervised Label-Proxy Alignment}

In the supervised setting, 
the availability of task labels \(Y\) enables direct supervision of masked partial modality combinations.
\modify{The} supervised \modify{objective uses} a label-conditioned target \modify{to incorporate task supervision, whereas the unsupervised objective uses label-agnostic consistency between masked and unmasked fused views}.
\model optimizes a label-proxy objective that requires each masked partial modality combination to align with the correct class proxy.
\modify{Each} masked fused representation is derived from a different combination \modify{of retained tokens across modalities. Requiring} label-relevant consistency across \modify{these partially observed token sequences encourages} the \modify{model to} capture synergistic interactions among modalities.

Given the fused representation \(Z\), we use a linear classifier parameterized by class proxies
\(W=\{w_j\}_{j=1}^{C}\), where \(C\) denotes the number of classes. 
For an input \({X}\), let \(z\) be the realization of \(Z\). The logit for class \(j\) is computed as \(w_j^\top z\), and the corresponding predictive probability is obtained using the softmax function: 
\begin{equation} 
        \modify{p_{\theta,W}}(Y=j\mid X)
        =
        \frac{\exp\left(w_j^\top \modify{z_\theta}(X)\right)}
        {\modify{\sum_{\ell=1}^{C}}\exp\left(\modify{w_\ell^\top z_\theta(X)}\right)} .
        \label{eq:fused_prediction}
\end{equation} 

Following Eq.~(\ref{eq:fused_prediction}), given the masked fused representation \(Z_{\mathrm{mask}}\), the label-proxy cross-entropy loss for a sample with \(Y=c\) is defined as
\begin{equation}
    \mathcal{L}_{\text{CE}} = -\log
    \frac{\exp(w_c^\top z_{\mathrm{mask}})}
    {\sum_{j=1}^{C}\exp(w_j^\top z_{\mathrm{mask}})}.
    \label{eq:finite_masked_ce}
\end{equation}
We again \modify{formulate} the \modify{objective over the mask distribution} to \modify{train on} diverse \modify{partial token combinations} across \modify{modalities}.
Accordingly, we define the supervised infinite masking objective \(\mathcal{L}_{\mathrm{InfMasking}}^{\mathrm{sup}}\) as the expected cross-entropy loss:
\begin{equation}
    \mathcal{L}_{\mathrm{InfMasking}}^{\mathrm{sup}}
    =
    \lim_{k\rightarrow \infty}\frac{1}{k}\sum_{i=1}^k
    \left[
        \mathcal{L}_{\text{CE}}^{(i)}
    \right]
    =
    \mathbb{E}_{\mathrm{mask}}
    \left[
        \mathcal{L}_{\text{CE}}
    \right],
    \label{eq:sup_infmasking}
\end{equation}
where \(\mathcal{L}_{\text{CE}}^{(i)}\) denotes the loss in Eq.~(\ref{eq:finite_masked_ce}) evaluated on the masked fused representation produced by the \(i\)-th sampled mask.
This objective instantiates Definition~\ref{def:infmasking} with the class-proxy set as the reference target.
Similarly, the above expectation cannot be evaluated exactly, and we approximate it with the same moment-based surrogate.
Building on the Gaussian approximation introduced above, 
\textit{i.e.}, \(z_{\mathrm{mask}} \sim \mathcal{N}(\boldsymbol{\mu}, \boldsymbol{\Sigma})\),
we obtain the following variational upper bound for the label-proxy cross-entropy loss.
The complete proof is provided in the supplementary material.

\begin{restatable}{lemma}{supInfMasking}
    Let \(\boldsymbol{\mu}\) and \(\boldsymbol{\Sigma}\) denote the mean vector and covariance matrix of the Gaussian approximation to \(z_{\mathrm{mask}}\), respectively, and let \(w_j\), \(j=1,2,\ldots,C\), denote the class proxy vectors in the final fully connected layer, where \(w_c\) corresponds to the correct class.
    Define \(\mathcal{J}=\mathbb{E}_{\mathrm{mask}}[\mathcal{L}_{\mathrm{CE}}]\). Then, the following upper bound holds:
    \begin{equation}
        \mathcal{J}
        \leq
        -w_c^\top \mu + \log \left( \sum_{j=1}^C \exp\left( w_j^\top \mu + \frac{1}{2} w_j^\top \Sigma w_j \right) \right).
    \end{equation}
    \label{lemma:sup_ce_gaussian_bound}
\end{restatable}
Lemma~\ref{lemma:sup_ce_gaussian_bound} \modify{gives} a tractable surrogate \modify{for} the label-proxy cross-entropy objective \modify{without enumerating all masks.}
\modify{Under} the \modify{Gaussian model}, the slack of this surrogate is \modify{controlled by the masking-induced covariance}; we formalize this property in Section~\ref{sec:tightness} and \modify{measure the observed surrogate gaps} in Section~\modify{\ref{sec:tightness_exp}}.
To preserve modality-specific discriminative information,
we follow the design of MMML~\cite{wu2024multimodal} and impose an auxiliary cross-entropy loss on each unimodal branch.
These auxiliary terms preserve \modify{each modality's} predictive contribution during joint training and \modify{help} the \modify{fusion network integrate} complementary modality-specific signals\modify{. The} label-proxy masking objective \modify{exposes the same network to partial modality combinations to promote} the learning of synergistic interactions.
Together with the standard cross-entropy loss on the complete fused representation, these terms yield the complete supervised objective:
\begin{equation}
    \begin{aligned}
    \mathcal{L}_{\mathrm{sup}}
    =
    &\left[
    -\log
    \frac{\exp(w_c^\top Z)}
    {\sum_{j=1}^{C}\exp(w_j^\top Z)}
    \right] \\
    &+\sum_{i=1}^{n}
    \left[
    -\log
    \frac{\exp(w_c^\top Z_i)}
    {\sum_{j=1}^{C}\exp(w_j^\top Z_i)}
    \right] \\
    &+\mathbb{E}_{\mathrm{mask}}
    \left[
    -\log
    \frac{\exp(w_c^\top Z_{\mathrm{mask}})}
    {\sum_{j=1}^{C}\exp(w_j^\top Z_{\mathrm{mask}})}
    \right].
    \end{aligned}
    \label{eq:full_sup_loss}
\end{equation}
The first term trains the complete fusion branch to predict the task label from all modalities. 
The second term preserves the predictive capacity of each unimodal branch. 
The final term enforces label-proxy consistency across partial modality combinations, thereby facilitating collaborative learning among modalities.
In practice, the masked expectation in this term is optimized through the variational upper bound in Lemma~\ref{lemma:sup_ce_gaussian_bound}, with \(\boldsymbol{\mu}\) and \(\boldsymbol{\Sigma}\) estimated from \(k\) sampled masks.
Note that despite the multiple training terms, the final classification uses only the fused representation of all modalities for prediction.
Although we present the supervised objective for classification, it extends directly to regression: with a linear regression head \(w\) and the squared loss, the masked objective admits the exact closed form \(\mathbb{E}_{\mathrm{mask}}[(w^\top Z_{\mathrm{mask}}-y)^2]=(w^\top\boldsymbol{\mu}-y)^2+w^\top\boldsymbol{\Sigma}w\), which requires no variational bound; our experiments focus on the classification setting.

\subsection{Tightness of the Variational Bounds}
\label{sec:tightness}

Both Lemma~\ref{lemma:unsup_inf} and Lemma~\ref{lemma:sup_ce_gaussian_bound} \modify{approximate} an intractable expectation over infinitely many masks with a closed-form surrogate under a Gaussian model of the masked fusion distribution. \modify{We examine the discrepancy between each} surrogate \modify{and} the true masked objective, and \modify{whether its characterization requires exact Gaussianity} or \modify{extends through} a directly measurable quantity \modify{to other masked-representation distributions.}

The two theorems below \modify{address} both \modify{issues} under the Gaussian model: each gap is (i) always non-negative and (ii) upper-bounded by a quadratic form of the masking-induced covariance $\Sigma$ (resp.\ $\Sigma'$) along the class-proxy (resp.\ anchor) directions, vanishing once masking induces no variation in the fused representation; both facts follow from two applications of Jensen's inequality. Remark~\ref{remark:beyond_gaussian} then delineates which parts of these guarantees survive beyond the Gaussian model\modify{. The supplementary tightness table reports masking-induced covariance statistics alongside observed surrogate gaps for} the \modify{tested configurations}.

\begin{restatable}[Tightness of the unsupervised bound]{theorem}{tightnessUnsup}
\label{prop:tightness_unsup}
Let $z'_{\mathrm{mask}}\sim\mathcal{N}(\mu',\Sigma')$ conditionally on the underlying input, with $\mu',\Sigma'$ the input-dependent conditional moments; the expectation $\mathbb{E}_{z'}$ below is over the joint distribution of the anchor and these moments. \modify{We further assume that this conditional covariance is uniformly bounded in spectral norm, that is, $\bar\lambda'_{\max}:=\operatorname{ess\,sup}_{x}\lambda_{\max}(\Sigma'(x))<\infty$; this assumption is used only for the last, spectral inequality below.} Let $\mathcal{L}'$ denote the lower bound of Lemma~\ref{lemma:unsup_inf}, and let $\mathcal{J}'=\mathbb{E}_{\mathrm{mask}}[\widehat{I}_{\mathrm{NCE}}(\modify{Z',Z'_{\mathrm{mask}}})]$ denote the true masked NCE objective that $\mathcal{L}'$ approximates. Then
\begin{equation}
   \resizebox{\linewidth}{!}{$\displaystyle
   0\;\le\;\mathcal{J}'-\mathcal{L}'
   \;\le\;
   \underbrace{\mathbb{E}_{z'}\!\left[\tfrac{1}{2\tau^2}z'^\top\Sigma'z'\right]}_{B'^{\mathrm{quad}}}
   \;\le\;
   \underbrace{\tfrac{1}{2\tau^2}\modify{\bar\lambda'_{\max}}\,\mathbb{E}_{z'}[\|z'\|_2^2]}_{B'^{\mathrm{theory}}}.
   $}
   \label{eq:tightness_unsup}
\end{equation}
In particular, when the fused views are $\ell_2$-normalized ($\|z'\|_2=1$) the slack is at most $\modify{\bar\lambda'_{\max}}/(2\tau^2)$, and both bounds vanish as $\modify{\bar\lambda'_{\max}}\to0$.
\end{restatable}

The supervised bound admits an analogous, symmetric guarantee.

\begin{restatable}[Tightness of the supervised bound]{theorem}{tightnessSup}
\label{prop:tightness}
Fix an input with correct class $c$, and let $z_{\mathrm{mask}}\sim\mathcal{N}(\mu,\Sigma)$, where $\mu$ and $\Sigma$ denote the conditional mean and covariance of the masked fused representation of this input. Let $W=[w_1,\dots,w_C]\in\mathbb{R}^{d\times C}$ collect the class proxies, and define $s_i=w_i^\top\mu$ and $a_i=\tfrac12 w_i^\top\Sigma w_i$ for $i=1,\dots,C$, where $a_i$ quantifies how much masking perturbs the $i$-th class logit. Let $\mathcal{U}$ denote the upper bound established in Lemma~\ref{lemma:sup_ce_gaussian_bound}, and let $\mathcal{J}=\mathbb{E}_{\mathrm{mask}}[\mathcal{L}_{\mathrm{CE}}]$ denote the true masked cross-entropy objective that $\mathcal{U}$ approximates. Then
\begin{equation}
   \resizebox{\linewidth}{!}{$\displaystyle
   0\;\le\;\mathcal{U}-\mathcal{J}
   \;\le\;
   \underbrace{\max_i a_i}_{B^{\mathrm{quad}}}
   \;\le\;
   \underbrace{\tfrac{1}{2}\lambda_{\max}(\Sigma)\max_i\|w_i\|_2^2}_{B^{\mathrm{theory}}}
   \;\le\;
   \tfrac{1}{2}\,\sigma_{\max}^2(W)\,\lambda_{\max}(\Sigma),
   $}
   \label{eq:tightness}
\end{equation}
where $\sigma_{\max}(W)$ is the largest singular value of $W$ and $\lambda_{\max}(\Sigma)$ the largest eigenvalue of $\Sigma$. In particular, $\mathcal{U}-\mathcal{J}\to0$ as $\lambda_{\max}(\Sigma)\to0$.
\end{restatable}

\modify{In both theorems, the quadratic bound is no larger than the subsequent spectral relaxation. The quadratic forms expose the dependence of the surrogate gap on masking-induced covariance, while the spectral relaxations summarize this dependence through eigenvalues and vector norms.}
\modify{The proofs also yield intermediate bounds. In the supervised setting, define $B^{\mathrm{LSE}}=\mathrm{LSE}(s+a)-\mathrm{LSE}(s)$, so that $\mathcal{U}-\mathcal{J}\le B^{\mathrm{LSE}}\le B^{\mathrm{quad}}\le B^{\mathrm{theory}}$. In the unsupervised setting, define $B'^{\mathrm{LSE}}=\mathbb{E}_{z'}[\log(e^{s'+a'}+c_{z'})-\log(e^{s'}+c_{z'})]$, where $s'=z'^\top\mu'/\tau$, $a'=z'^\top\Sigma'z'/(2\tau^2)$ and $c_{z'}=\sum_{z'_{\mathrm{neg}}}\exp(z'^\top z'_{\mathrm{neg}}/\tau)$. Then $\mathcal{J}'-\mathcal{L}'\le B'^{\mathrm{LSE}}\le B'^{\mathrm{quad}}\le B'^{\mathrm{theory}}$; the last inequality requires the uniform spectral condition in Theorem}~\modify{\ref{prop:tightness_unsup}. The theorem statements use the quadratic bounds to make their dependence on masking-induced covariance explicit. These inequalities use population conditional moments.} 

\begin{remark}[Scope of the guarantees beyond the Gaussian model]
\label{remark:beyond_gaussian}
Theorems~\ref{prop:tightness_unsup} and~\ref{prop:tightness} are stated under the Gaussian model; their conclusions split into two parts with different ranges of validity.
(i) The \emph{magnitude} of the gap is controlled distribution-free: with $\mu,\Sigma$ (resp.\ $\mu',\Sigma'$) taken as the true mean and covariance of the masked fused representations under an arbitrary mask distribution, the upper estimates $\mathcal{U}-\mathcal{J}\le B^{\mathrm{quad}}$ and $\mathcal{J}'-\mathcal{L}'\le B'^{\mathrm{quad}}$ invoke only the convexity of the log-sum-exp term, and matching moment-only lower estimates (Proposition~1 in the supplementary material) show that $|\mathcal{U}-\mathcal{J}|$ and $|\mathcal{J}'-\mathcal{L}'|$ vanish with the masking-induced covariance.
(ii) The \emph{sign} of the gap\modify{, namely} the statements $\mathcal{U}\ge\mathcal{J}$ and $\mathcal{J}'\ge\mathcal{L}'$, is \emph{not} distribution-free\modify{. A sufficient condition is} sub-Gaussianity with variance proxy $\Sigma$ along the class-proxy directions, $\mathbb{E}_{\mathrm{mask}}[\exp(w_i^\top(z_{\mathrm{mask}}-\mu))]\le\exp(\tfrac12 w_i^\top\Sigma w_i)$, and analogously along each anchor $z'$\modify{. This condition is} weaker than exact Gaussianity, but \modify{remains an} assumption on the masked representations.
\modify{The} non-negative \modify{mean} Monte-Carlo gaps \modify{reported} in Section~\modify{\ref{sec:tightness_exp} do not establish this directional condition}.
\end{remark}

\textbf{\modify{Scope and boundary conditions of the surrogate.}}
\modify{Small masking-induced covariance along the relevant class-proxy or anchor directions gives a small absolute approximation error, whose relative size depends on the objective scale. For a fixed model and input, this covariance vanishes at both \(\rho=0\) and \(\rho=1\); the sweeps in Section}~\modify{\ref{sec:tightness_exp} evaluate intermediate ratios and report larger and more variable gaps on MUStARD. The sampled mask count \(k\) affects empirical moment estimates, and the empirical covariance has rank at most \(k-1\). The surrogate evaluates scalar covariance projections without matrix inversion, so they remain well-defined at low rank; their accuracy depends on finite-sample variability. Absolute-error control with true conditional moments also applies when the sufficient directional condition fails. Gap signs and downstream predictive effects are distinct from this error control.}

The complete proofs are provided in the supplementary material.
\modify{For $k$ masked representations $z_1,\ldots,z_k$}, \modify{we use $\widehat\mu=k^{-1}\sum_jz_j$ and $\widehat\Sigma=k^{-1}\sum_j(z_j-\widehat\mu)(z_j-\widehat\mu)^\top$, with analogous primed estimators in the unsupervised setting}. \modify{These} are the \modify{moments} of the \modify{equal-weight empirical mask distribution. The $1/k$ covariance is a plug-in estimate} of the \modify{population covariance, without a Bessel correction. The} supplementary \modify{ablations report predictive performance across mask counts; Section}~\modify{\ref{sec:tightness_exp} reports finite-mask surrogate gaps}.

\section{Experiments in the Unsupervised Setting}
\label{sec:unsup_exp}

\textbf{Evaluation Protocol.}
To evaluate unsupervised representation learning, we adopt a linear probing protocol that trains a linear classifier (or regressor) on top of the frozen feature extractors after pre-training, so that downstream performance primarily reflects the quality of the learned representations.
We first use the synthetic Trifeature dataset~\cite{hermann2020shapes} to examine the ability of \model to capture different multimodal information components, \textit{i.e.}, redundancy, uniqueness, and synergy, under controlled generative factors.
Furthermore, we evaluate the transfer performance of \model on two real-world benchmarks, MultiBench~\cite{liang2021multibench} and MM-IMDb~\cite{arevalo2017gated}, which cover realistic modality combinations, label structures, and prediction objectives.
Detailed dataset descriptions and experimental settings are provided in the supplementary material.
\modify{The supplementary material also contains the masking-ratio and mask-count ablations, test-time robustness results, and surrogate-gap measurements.}

\subsection{Synthetic Trifeature Benchmark}

Following the Trifeature protocol adopted by CoMM~\cite{dufumier2024align}, we construct paired-image tasks that isolate different multimodal interactions as redundancy, uniqueness, and synergy. 
Each image is characterized by one of ten shapes, one of ten textures, and one of ten colors, resulting in 1,000 unique combinations. 
Among them, 800 combinations are used for training and the remaining 200 are reserved for testing.
For redundancy and uniqueness, shape is the shared factor across paired images and the texture of one image is a modality-specific factor; the two tasks evaluate whether the representation encodes shared shape and image-specific texture, respectively, both as ten-way classification with a 10\% random-guessing baseline.
For synergy, \modify{the CoMM protocol}~\modify{\cite{dufumier2024align} uses} a training-set mapping \(\mathcal{A}\) between texture and color attributes (\textit{e.g.}, stripes map to red). \modify{Unlabeled representation pre-training uses pairs that satisfy this mapping; downstream linear-probe training uses pairs that can satisfy or violate it.} The binary \modify{probe} label is \(Y=\mathbf{1}\{(\mathrm{texture}(X_1),\mathrm{color}(X_2))\in\mathcal{A}\}\)\modify{. Uniform random guessing between the two classes gives} 50\% \modify{accuracy}, \modify{but this does not imply balanced class frequencies. The} synergy task \modify{tests} whether \modify{a representation of} the paired \modify{images supports prediction of their texture--color correspondence}.
The baselines include ``Cross''~\cite{radford2021learning}, ``Cross+Self''~\cite{yuan2021multimodal}, FactorCL~\cite{liang2023factorized}, MAE~\cite{he2022masked}, and CoMM~\cite{dufumier2024align}.
\modify{As shown in Table}~\modify{\ref{tab:trifeature_results},} 
\modify{all models achieve near-perfect accuracy on the redundancy task, indicating that shared shape information can be effectively captured under the multiview redundancy assumption.}
\modify{However, their performance differs on the uniqueness and synergy tasks.}
\modify{The ``Cross'' baseline remains close to chance on both tasks, suggesting that cross-modal alignment alone primarily captures shared factors and does not preserve modality-specific information or synergistic interactions.}
\modify{Although ``Cross+Self'' and MAE substantially improve uniqueness, their synergy scores remain close to the 50\% random-guessing baseline, indicating that they still struggle to capture compositional information arising from modality interactions.}
\modify{FactorCL exhibits a similar limitation in this controlled setting.}
\modify{CoMM reaches 71.4\% on the synergy task.}
\modify{\model further improves uniqueness to 90.6\% and synergy to 77.0\%,}
\modify{exceeding CoMM by 3.8 and 5.6 percentage points, respectively, while maintaining near-perfect redundancy accuracy.}
\modify{The gains on uniqueness and synergy show that \model retains modality-specific information alongside joint evidence in this controlled task.}

\begin{table}[t]
    \centering
    \caption{Linear-probing accuracy (\%) on the synthetic Trifeature dataset. Redundancy measures shared shape recognition, uniqueness measures texture recognition, and synergy measures texture--color composition. 
    $^\clubsuit$ denotes that results are from~\cite{dufumier2024align}.}
    \label{tab:trifeature_results}
    \begin{tabular}{lccc}
        \toprule
        \textit{Model} & \textit{Redundancy$\uparrow$} & \textit{Uniqueness$\uparrow$} & \textit{Synergy$\uparrow$} \\
        \midrule
        Cross$^\clubsuit$~\cite{radford2021learning} & $\textbf{100.0}$ & $11.6$ & $50.0$ \\
        Cross+Self$^\clubsuit$~\cite{yuan2021multimodal} & $99.7$ & $86.9$ & $50.0$ \\
        FactorCL$^\clubsuit$~\cite{liang2023factorized} & $99.8$ & $62.5$ & $46.5$ \\
        MAE~\cite{he2022masked} & $99.8_{\pm0.11}$ & $82.4_{\pm3.09}$ & $50.1_{\pm0.24}$ \\
        CoMM~\cite{dufumier2024align} & $99.9_{\pm0.06}$ & $86.8_{\pm2.99}$ & $71.4_{\pm3.47}$ \\
        \model (ours) & $99.9_{\pm0.09}$ & $\textbf{90.6}_{\pm2.31}$ & $\textbf{77.0}_{\pm4.22}$ \\
        \bottomrule
    \end{tabular}
\end{table}

\subsection{Real-World Benchmark Evaluation}

We further evaluate \model on real-world multimodal benchmarks from MultiBench~\cite{liang2021multibench} and MM-IMDb~\cite{arevalo2017gated} to assess its effectiveness beyond controlled synthetic settings.
These benchmarks cover healthcare, sentiment analysis, robotics, and movie genre classification, \modify{with} heterogeneous modality combinations such as vision, text, audio, force, and proprioception\modify{. They test the learned representations on} both regression and classification objectives.

\paragraph{Bimodal MultiBench results}
\label{para:multibench_results}
Following the preprocessing procedure of previous work~\cite{liang2023factorized,dufumier2024align}, we use the same encoders, modality configurations, and encoded inputs to conduct our evaluation.
The comparison includes ``Cross''~\cite{radford2021learning}, ``Cross+Self''~\cite{yuan2021multimodal}, FactorCL~\cite{liang2023factorized}, and CoMM~\cite{dufumier2024align}\modify{, as well as MultiMAE-inspired}~\modify{\cite{bachmann2022multimae} and DCP-inspired}~\modify{\cite{lin2022dual} masked multimodal self-supervised baselines}. \modify{MultiMAE reconstructs masked tokens across modalities, whereas DCP combines within-view reconstruction, cross-view contrastive learning, and dual prediction across views. For these fully observed bimodal datasets, our DCP-inspired adaptation treats each modality as a view and applies random token masking during training rather than assuming naturally missing views. Both adaptations use the shared encoders and augmentations and the same training budget as \model.}

As shown in Table~\ref{tab:Multibench_results}, \model achieves the best overall performance across heterogeneous task types, including both regression and classification, and across diverse modality combinations.
For classification, \model obtains the highest accuracy on all four datasets, improving over CoMM by 1.7, 5.3, 1.0, and 2.4 percentage points on MIMIC, MOSI, UR-FUNNY, and MUStARD, respectively, and increasing the average classification accuracy from 64.45\% to 67.05\%.
For the Vision\&Touch end-effector regression task, \model reduces the MSE from the second-best $7.56\times10^{-4}$ to $4.23\times10^{-4}$, indicating a substantial improvement in continuous prediction.
Because \modify{the} downstream heads are linear, these gains \modify{indicate} that the \modify{learned} representations \modify{support both classification and regression more effectively.}
\modify{The MultiMAE-inspired and DCP-inspired adaptations average $65.10\%$ and $65.38\%$ over the four classification datasets, against $67.05\%$ for \model; both score lower on each dataset. Both improve on CoMM on MIMIC, and the DCP-inspired variant also improves on CoMM on MOSI. Because the adaptations differ} from \model \modify{in their objectives and auxiliary modules, these results compare predictive performance without isolating alignment across masks. Implementation details} are \modify{in the supplementary material}.

\begin{table*}[t]
    \centering
    \caption{Linear-probing performance on bimodal MultiBench datasets in the unsupervised setting. 
    Vision\&Touch end-effector prediction reports MSE ($\times 10^{-4}$; lower is better), whereas MIMIC, MOSI, UR-FUNNY, and MUStARD report top-1 accuracy (\%). 
    $^\clubsuit$ denotes results reported in~\cite{liang2023factorized}. 
    $^*$ denotes the average over classification tasks only.
    }
    \label{tab:Multibench_results}
    \begin{tabular}{lcccccc}
        \toprule
        \multirow{2}{*}{\textit{Model}} & \multicolumn{1}{c}{\textit{Regression}} & \multicolumn{5}{c}{\textit{Classification}} \\
        \cmidrule(lr){2-2} \cmidrule(lr){3-7}
        & \textit{V\&T EE}$\downarrow$ & \textit{MIMIC}$\uparrow$ & \textit{MOSI}$\uparrow$ & \textit{UR-FUNNY}$\uparrow$ & \textit{MUStARD}$\uparrow$ & \textbf{Average}$^{*}\uparrow$\\
        \midrule
        $\text{Cross}^\clubsuit$~\cite{radford2021learning} & $33.09_{\pm3.67}$ & $66.7_{\pm0.1}$ & $47.8_{\pm1.8}$ & $50.1_{\pm1.9}$ & $53.5_{\pm2.9}$ & $54.\modify{53}$ \\
        $\text{Cross+Self}^\clubsuit$~\cite{yuan2021multimodal} & $7.56_{\pm0.31}$ & $65.49_{\pm0.0}$ & $49.0_{\pm1.1}$ & $59.9_{\pm0.9}$ & $53.9_{\pm4.0}$ & $57.07$ \\
        $\text{FactorCL}^\clubsuit$~\cite{liang2023factorized} & $10.82_{\pm0.56}$ & $67.3_{\pm0.0}$ & $51.2_{\pm1.6}$ & $60.5_{\pm0.8}$ & $55.8_{\pm0.9}$ & $58.7$ \\
        \modify{MultiMAE-inspired}~\modify{\cite{bachmann2022multimae}} & \modify{$25.19_{\pm12.65}$} & \modify{$67.8_{\pm0.3}$} & \modify{$64.2_{\pm3.0}$} & \modify{$63.5_{\pm0.9}$} & \modify{$64.9_{\pm4.3}$} & \modify{$65.10$} \\
        \modify{DCP-inspired}~\modify{\cite{lin2022dual}} & \modify{$8.07_{\pm1.86}$} & \modify{$67.7_{\pm0.4}$} & \modify{$66.1_{\pm2.6}$} & \modify{$64.1_{\pm1.0}$} & \modify{$63.6_{\pm2.9}$} & \modify{$65.38$} \\
        CoMM~\cite{dufumier2024align} & $7.96_{\pm2.13}$ & $66.4_{\pm0.41}$ & $63.7_{\pm2.5}$ & $63.3_{\pm0.51}$ & $64.4_{\pm1.1}$ & $64.45$ \\
        \model (ours) & $\textbf{4.23}_{\pm0.37}$ & $\textbf{68.1}_{\pm0.42}$ & $\textbf{69.0}_{\pm1.2}$ & $\textbf{64.3}_{\pm0.9}$ & $\textbf{66.8}_{\pm2.5}$ & $\textbf{67.05}$ \\
    \bottomrule
    \end{tabular}
\end{table*}

\paragraph{Trimodal MultiBench results}
We further evaluate the scalability of \model to more than two modalities on the Vision\&Touch (with visual, force, and proprioceptive modalities) contact prediction task and the UR-FUNNY (with visual, text, and audio modalities) humor detection task.
CMC~\cite{tian2020cmc} and CoMM~\cite{dufumier2024align} are included as trimodal baselines.

\modify{As shown in Table}~\modify{\ref{tab:tri_results}, adding the third modality substantially improves both CoMM and \model compared with the bimodal setting.} 
\modify{For CoMM, accuracy increases from 87.0\% to 94.1\% on Vision\&Touch and from 63.3\% to 64.8\% on UR-FUNNY.} 
\modify{\model follows the same pattern, improving from 88.5\% to 94.1\% and from 64.3\% to 65.6\%, respectively.} 
\modify{On UR-FUNNY, \model achieves the highest accuracy and exceeds the three-modality CoMM result by 0.8 percentage points.}

\begin{table}[t]
    \centering
    \caption{Linear-probing top-1 accuracy (\%) on the Vision\&Touch contact prediction and UR-FUNNY humor detection tasks in the unsupervised setting.
    \#Mod. denotes the number of input modalities. 
    $^\clubsuit$ denotes results are from~\cite{dufumier2024align}.}
    \label{tab:tri_results}
    \begin{tabular}{lccc}
        \toprule
        \textit{Model} & \textit{\#Mod.} & \textit{V\&T CP$\uparrow$} & \textit{UR-FUNNY$\uparrow$} \\
        \midrule
        Cross & 2 & $86.3_{\pm0.25}$ & $50.1^\clubsuit$ \\
        Cross+Self & 2 & $87.6_{\pm0.26}$ & $59.9^\clubsuit$ \\
        CoMM & 2 & $87.0_{\pm1.77}$ & $63.3_{\pm0.51}$ \\
        \model (ours) & 2 & $88.5_{\pm0.33}$ & $64.3_{\pm0.9}$ \\
        \midrule
        CMC$^\clubsuit$~\cite{tian2020cmc} & 3 & $94.1$ & $59.2$ \\
        CoMM & 3 & $\textbf{94.1}_{\pm0.17}$ & $64.8_{\pm1.13}$ \\
        \model (ours) & 3 & $\textbf{94.1}_{\pm0.09}$ & $\textbf{65.6}_{\pm1.15}$ \\
        \bottomrule
    \end{tabular}
\end{table}

\paragraph{Bimodal results on MM-IMDb}
\begin{table}[t]
    \centering
    \caption{Linear-probing F1-scores (\%) on the MM-IMDb multi-label genre classification task in the unsupervised setting.
    V and L denote visual and language modalities, respectively. 
    $^\triangle$ indicates additional training on unlabeled data, 
    and $^\clubsuit$ denotes results are from~\cite{dufumier2024align}.}
    \label{tab:imdb_results}
    \begin{tabular}{lccc}
        \toprule
        \textit{Model} & \textit{Modalities} & \textit{Weighted F1$\uparrow$} & \textit{Macro F1$\uparrow$} \\
        \midrule
        SimCLR$^{\clubsuit\triangle}$~\cite{chen2020simple} & V & $40.35_{\pm0.23}$ & $27.99_{\pm0.33}$ \\
        \multirow{3}{*}{CLIP$^\clubsuit$~\cite{radford2021learning}} & V & $51.5$ & $40.8$ \\ & L & $51.0$ & $43.0$ \\
        & V+L & $58.9$ & $50.9$ \\
        SLIP$^{\clubsuit\triangle}$~\cite{mu2022slip} & V+L & $56.54_{\pm0.19}$ & $47.35_{\pm0.27}$ \\
        CLIP$^{\clubsuit\triangle}$~\cite{radford2021learning} & V+L & $54.49_{\pm0.19}$ & $44.94_{\pm0.30}$ \\
        CoMM & V+L & $61.29_{\pm0.73}$ & $53.79_{\pm0.22}$ \\
        \model (ours) & V+L & $\textbf{62.60}_{\pm0.26}$ & $\textbf{55.93}_{\pm0.19}$ \\
        \bottomrule
    \end{tabular}
\end{table}
Multimodal IMDb (MM-IMDb)~\cite{arevalo2017gated} is a multi-label movie genre classification dataset with poster images and plot descriptions, where frequent genres dominate the label distribution and the two modalities often provide semantically different evidence, making effective fusion critical.
We compare \model with SimCLR~\cite{chen2020simple}, CLIP~\cite{radford2021learning} evaluated under three modality configurations (V, L, and V+L), CLIP and SLIP~\cite{mu2022slip} variants further trained on unlabeled data, and CoMM~\cite{dufumier2024align}.

Table~\ref{tab:imdb_results} reports weighted and macro F1-scores for all models. 
Multimodal CLIP outperforms its single-modality variants, confirming that the two modalities provide complementary genre evidence. 
CoMM further improves both metrics, and \model achieves the strongest result with 62.60\% weighted F1 and 55.93\% macro F1. 
Compared with CoMM, the gains are 1.31 and 2.14 percentage points, respectively. 
The larger macro F1 \modify{gain} is \modify{relevant to MM-IMDb's imbalanced labels:} macro averaging gives equal weight to rare and frequent genres and is therefore more sensitive to under-represented labels.
In addition, the publicly released CLIP model outperforms the CLIP variant further trained on unlabeled MM-IMDb data, suggesting that learning redundant image-text agreement alone is insufficient to improve performance on this imbalanced multi-label benchmark.
\modify{On MM-IMDb, \model's gains over CoMM support its effectiveness for multi-label genre prediction with imbalanced labels.}

\subsection{Ablation Studies}
\label{unsup_ablation}

We conduct unsupervised ablation studies on the bimodal Trifeature benchmark to isolate the contributions of the objective terms and masking hyperparameters\modify{, and, for the comparison between estimators of the masked expectation, on the four bimodal MultiBench classification benchmarks, whose accuracies are directly comparable with Table}~\modify{\ref{tab:Multibench_results}}. 
All ablations follow the same linear-probing evaluation protocol as above. 

\begin{table*}[t]
    \centering
    \caption{
        Ablation of the unsupervised objective (Eq.~(\ref{eq:full_unsup_loss})): linear-probing accuracy (\%) on Trifeature. \(\lambda_1\), \(\lambda_2\), and \(\lambda_3\) indicate whether the masked fusion, unimodal alignment, and fused contrastive terms are included (1) or excluded (0).
    }
    \label{tab:loss_weights}
    \begin{tabular}{cccccccc}
        \toprule
        \(\lambda_1\) & \(\lambda_2\) & \(\lambda_3\) & \textit{R} & \textit{$U_1$} & \textit{$U_2$} & \textit{S} & \textit{Average} \\
        \midrule
        0 & 0 & 1 & $95.8_{\pm1.91}$ & $85.9_{\pm2.11}$ & $83.8_{\pm2.97}$ & $58.7_{\pm7.11}$ & $81.1$ \\
        0 & 1 & 1 & $\textbf{99.9}_{\pm0.06}$ & $87.1_{\pm3.31}$ & $86.5_{\pm2.60}$ & $71.4_{\pm3.47}$ & $86.2$\\
        1 & 1 & 0 & $\textbf{99.9}_{\pm0.08}$ & $\textbf{90.7}_{\pm2.10}$ & $\textbf{91.4}_{\pm3.03}$ & $69.2_{\pm6.20}$ & $87.8$ \\
        1 & 1 & 1 & $\textbf{99.9}_{\pm0.09}$ & $90.3_{\pm1.52}$ & $90.8_{\pm2.88}$ & $\textbf{77.0}_{\pm4.22}$ & $\textbf{89.5}$ \\
        \bottomrule
        \end{tabular}
\end{table*}

\paragraph{Loss terms}
Table~\ref{tab:loss_weights} compares different combinations of the \model loss, unimodal loss, and fused contrastive loss. 
The full objective (\(\lambda_1=\lambda_2=\lambda_3=1\), \model) yields the best average score and the highest synergy accuracy of 77.0\%. 
Removing the \model term, which corresponds to the CoMM objective, lowers synergy to 71.4\%, while using only the fused contrastive term decreases it further to 58.7\%. 
Conversely, removing the fused contrastive term slightly improves uniqueness within a fixed training budget but reduces synergy to 69.2\%, reflecting its role in learning multiple information components~\cite{dufumier2024align}.
These trends indicate that the three objective terms play complementary roles: 
unimodal losses preserve modality-specific factors, 
fused contrastive alignment stabilizes shared representations, 
and masked fusion introduces additional view diversity that facilitates the learning of cross-modal interaction patterns.

\paragraph{Number of masked views} 
Increasing the number of sampled masked views yields a more accurate finite-sample approximation to the expectation over masks in Eq.~(\ref{eq:mask_loss}), at the expense of additional training cost.
\modify{On Trifeature, synergy} accuracy \modify{rises from 2 to 8} masked views and \modify{falls at} 10 views \modify{in} the supplementary \modify{figure}.
\modify{The Trifeature result favors} a moderate number of masked views \modify{among} the \modify{values tested}.

\paragraph{Masking ratio} 
The effect of the masking ratio on synergy learning is reported in the supplementary material.
At low masking ratios \((\leq 50\%)\), each masked view preserves most of the original evidence, 
providing only weak perturbations and therefore yielding limited gains in synergy accuracy. 
In contrast, higher masking ratios remove a larger fraction of the input features, 
encouraging the fusion module to infer cross-modal interactions from complementary cues that remain available across partial observations.

\paragraph{\modify{RandomMask versus \model}}
\modify{Table}~\modify{\ref{tab:masking_ablation} compares CoMM without masking, RandomMask, and \model. RandomMask randomly masks encoded feature tokens before fusion and directly averages the masked-view InfoNCE losses over $k$ sampled views to estimate Eq.}~\modify{(\ref{eq:mask_loss}). \model uses empirical moments of $k$ masked representations in the variational lower bound of Lemma}~\modify{\ref{lemma:unsup_inf}. Both reuse encoder features across masked fusion passes and include a masked-view loss.} 
\modify{The four-dataset average linear-probing accuracies are $64.45\%$ without masking, $65.57\%$ with RandomMask, and $67.05\%$ with \model. \model has a higher reported mean on all four datasets. RandomMask improves average accuracy over CoMM by $1.12$ percentage points, but its dataset-level effects are mixed: it improves over CoMM on MIMIC, MOSI, and UR-FUNNY and decreases accuracy on MUStARD. These results summarize predictive performance under the reported training configurations.}

\begin{table*}[t]
    \centering
    \caption{
        \modify{RandomMask and \model in the unsupervised setting (linear-probing accuracy, \%). Average is taken over the four datasets.}
    }
    \label{tab:masking_ablation}
    
    \begin{tabular}{llccccc}
        \toprule
        & \textit{\modify{Variant}} & \textit{\modify{MIMIC}} & \textit{\modify{MOSI}} & \textit{\modify{UR-FUNNY}} & \textit{\modify{MUStARD}} & \textbf{\modify{Average}} \\
        \midrule
        \modify{(a)} & \modify{CoMM objective, no masking} & \modify{$66.4_{\pm0.41}$} & \modify{$63.7_{\pm2.5}$} & \modify{$63.3_{\pm0.51}$} & \modify{$64.4_{\pm1.1}$} & \modify{$64.45$} \\
        \modify{(b)} & \modify{RandomMask ($k$ masks, no variational bound)} & \modify{$67.7_{\pm0.4}$} & \modify{$67.8_{\pm1.4}$} & \modify{$63.5_{\pm0.3}$} & \modify{$63.26_{\pm4.10}$} & \modify{$65.57$} \\
        \modify{(c)} & \modify{\model (variational bound, $k$ masks)} & \modify{$\textbf{68.1}_{\pm0.42}$} & \modify{$\textbf{69.0}_{\pm1.2}$} & \modify{$\textbf{64.3}_{\pm0.9}$} & \modify{$\textbf{66.8}_{\pm2.5}$} & \modify{$\textbf{67.05}$} \\
        \bottomrule
    \end{tabular}
\end{table*}

\section{Experiments in the Supervised Setting}
\label{sec:supervised_exp}

\textbf{Evaluation Protocol.}
For supervised evaluation, 
we train \model with the complete three-term objective in Eq.~(\ref{eq:full_sup_loss}), comprising cross-entropy on the fused representation, auxiliary unimodal cross-entropy losses, and the label-proxy masking objective in Eq.~(\ref{eq:sup_infmasking}),
and assess its performance on multimodal prediction tasks.
At inference time, predictions are made exclusively from the fused representation of all available modalities, 
followed by a softmax classifier.
We evaluate \model on benchmarks spanning diverse domains and modality configurations, 
including sentiment analysis (MOSI~\cite{zadeh2016multimodal} and MOSEI~\cite{zadeh2018multimodal}),
healthcare prediction (MIMIC~\cite{johnson2016mimic}) 
and
hate content detection (MAMI~\cite{fersini2022semeval} and MUTE~\cite{hossain2022mute}).
For each dataset, we compare \model against unimodal baselines and representative multimodal baselines,
including Vanilla MultiTrans (Vanilla MT) with attention fusion,
MulT~\cite{tsai2019multimodal}, MBT~\cite{nagrani2021attention}, JMT~\cite{waligora2024joint}, and MMML~\cite{wu2024multimodal}.
\modify{For each dataset}, \modify{the compared} methods use \modify{the same} modality-specific encoders and encoded inputs\modify{. The fusion-architecture baselines vary their fusion components, whereas Fusion only, Vanilla MT,} and \modify{\model use} the \modify{same} fusion \modify{backbone with different training losses}; \modify{the adapted MultiMAE- and DCP-inspired baselines additionally use auxiliary decoders or autoencoders during training.} \model introduces no additional learnable parameters over \modify{Vanilla MT}, \modify{and} its \modify{reported accuracy is higher on} the \modify{evaluated benchmarks. The supplementary material compares} parameter counts, FLOPs, \modify{MACs,} and \modify{peak GPU memory on MAMI, including the adapted masked-learning baselines. Parameter counts refer to the number of parameter elements, independently of storage precision. Adding \model to Vanilla MT preserves the $173.7$M parameters and increases peak GPU memory from $5.29$ to $5.38$ GiB ($0.09$ GiB, approximately $1.7\%$), while accuracy increases from $63.4\%$ to $69.3\%$} in \modify{this comparison}.
Additionally, dataset descriptions and training details are provided in the supplementary material.

\subsection{Real-World Benchmark Evaluation}

\begin{table*}[t]
    \centering
    \caption{Supervised performance (\%) on five multimodal benchmarks; T/V/A/TS/Tab denote unimodal text, visual, audio, time-series, and tabular baselines.
    $-$ denotes a result not reported and $\mathtt{\_}$ denotes the best performance among baseline methods.
    }
    \label{tab:supervised_benchmark_results}
    \begin{tabular}{lcccccc}
        \toprule
        \textit{Model} & \textit{MOSI} & \textit{MOSEI} & \textit{MIMIC} & \textit{MAMI} & \textit{MUTE} & \textbf{Average} \\
        \midrule
        Unimodal & \makecell{T: $72.8_{\pm0.83}$\\V: $57.6_{\pm2.12}$\\A: $51.7_{\pm2.31}$} &
        \makecell{T: $72.0_{\pm0.18}$\\V: $59.0_{\pm0.46}$\\A: $57.3_{\pm0.27}$} &
        \makecell{TS: $61.2_{\pm0.92}$\\Tab: $55.2_{\pm1.15}$} & 
        \makecell{T: $58.5_{\pm6.39}$\\V: $61.8_{\pm0.40}$} & 
        \makecell{T: $63.1_{\pm1.23}$\\V: $64.3_{\pm1.47}$} & -- \\
        \cmidrule(lr){1-7}
        \modify{Fusion only} & $\modify{71}.\modify{1_{\pm2.78}}$ &
                    $72.\modify{8_{\pm0.4}}$ &
                    $68.\modify{0_{\pm0.41}}$ &
                    $\modify{64}.\modify{7_{\pm4.61}}$ &
                    $64.\modify{3_{\pm0.97}}$ &
                    $68.\modify{2}$ \\
        Vanilla MT & $72.9_{\pm1.26}$ & 
                    $72.2_{\pm1.17}$ & 
                    $68.1_{\pm0.34}$ & 
                    $63.4_{\pm3.44}$ & 
                    $64.1_{\pm2.34}$ & 
                    $68.1$ \\
        MulT~\cite{tsai2019multimodal} & $73.1_{\pm1.54}$ & 
                    $72.4_{\pm0.21}$ & 
                    $67.6_{\pm0.40}$ & 
                    $\modify{65.6_{\pm0.99}}$ & 
                    $64.2_{\pm1.50}$ & 
                    $68.6$ \\
        MBT~\cite{nagrani2021attention} & $\underline{73.8}_{\pm0.34}$ & 
                    $\underline{73.4}_{\pm0.22}$ & 
                    $\underline{68.9}_{\pm0.19}$ & 
                    $62.8_{\pm1.91}$ & 
                    $\underline{67.5}_{\pm1.41}$ & 
                    $\modify{69.3}$ \\
        JMT~\cite{waligora2024joint} & $73.7_{\pm1.07}$ & 
                    $72.3_{\pm0.48}$ & 
                    $68.7_{\pm0.24}$ & 
                    $60.3_{\pm1.40}$ & 
                    $65.4_{\pm1.05}$ & 
                    $68.1$ \\
        MMML~\cite{wu2024multimodal} & $72.9_{\pm1.09}$ & 
                    $73.0_{\pm0.45}$ & 
                    $68.3_{\pm0.17}$ & 
                    $64.0_{\pm0.92}$ & 
                    $64.0_{\pm1.41}$ & 
                    $68.4$ \\
        \cmidrule(lr){1-7}
        \modify{MultiMAE-inspired}~\modify{\cite{bachmann2022multimae}} & \modify{$73.5_{\pm1.6}$} &
                    \modify{$72.4_{\pm0.8}$} &
                    \modify{$68.8_{\pm0.1}$} &
                    \modify{$66.1_{\pm0.6}$} &
                    \modify{$66.0_{\pm0.3}$} &
                    \modify{$69.4$} \\
        \modify{DCP-inspired}~\modify{\cite{lin2022dual}} & \modify{$73.4_{\pm0.9}$} &
                    \modify{$72.7_{\pm0.7}$} &
                    \modify{$68.7_{\pm0.1}$} &
                    \modify{$\underline{67.7}_{\pm2.5}$} &
                    \modify{$66.7_{\pm1.3}$} &
                    \modify{$\underline{69.8}$} \\
        \cmidrule(lr){1-7}
        Vanilla MT + \model & $\textbf{75.9}_{\pm0.89}$ & 
                    $\textbf{73.8}_{\pm0.25}$ & 
                    $\textbf{69.4}_{\pm0.06}$ & 
                    $\textbf{69.3}_{\pm1.60}$ & 
                    $\textbf{70.0}_{\pm2.87}$ & 
                    $\textbf{71.7}$ \\
        \bottomrule
    \end{tabular}
\end{table*}

Table~\ref{tab:supervised_benchmark_results} reports the supervised performance of \model and competing baselines across domain-diverse multimodal benchmark datasets.
\modify{Throughout the supervised experiments, ``Fusion only'' denotes the fused cross-entropy loss alone, ``Vanilla MT'' denotes that loss together with the auxiliary unimodal cross-entropy losses, and \model adds the label-proxy masking objective to the latter, so that the three rows correspond to the successive terms of Eq.}~\modify{(\ref{eq:full_sup_loss}). The MultiMAE-inspired and DCP-inspired rows are the masked multimodal self-supervised objectives of Section}~\modify{\ref{sec:unsup_exp} adapted to this supervised setting with the same encoders and training budget; they are the strongest baselines here, the DCP-inspired variant attaining the highest average accuracy and the best result on MAMI.}
Most multimodal fusion methods improve upon the strongest unimodal branch\modify{. On MUTE}, however, several fusion baselines remain close to or below the visual unimodal baseline\modify{, so adding} a fusion module \modify{alone does not} guarantee a \modify{gain}.

In contrast, \model achieves consistent improvements over the strongest unimodal baseline across all datasets 
and improves \modify{by 1.8 percentage points on average} over the \modify{DCP-inspired adaptation, which attains the highest average accuracy among the baselines; when the strongest} baseline \modify{is instead selected separately on each dataset, the margin is 1}.4 percentage points.
Furthermore, \model consistently improves upon its corresponding supervised backbone across all datasets.
Compared with the Vanilla MT backbone, \model improves MOSI, MOSEI, MIMIC, MAMI, and MUTE by 3.0, 1.6, 1.3, 5.9, and 5.9 percentage points, respectively,
raising the average classification accuracy from 68.1\% to 71.7\%.
Notably, the largest improvements are observed on MAMI and MUTE, 
two hate content detection datasets in which the label often depends on the joint interpretation of textual and visual cues rather than either modality alone.
\modify{Within the shared supervised fusion backbone, these} consistent gains \modify{over Vanilla MT support} the \modify{effectiveness of} label-proxy \modify{masking}.

\subsection{\modify{Ablation Studies}}
\label{sec:sup_ablation}

\modify{To assess \model's effectiveness in promoting modality collaboration} in the supervised setting\modify{, we examine the individual terms of Eq.}~\modify{(\ref{eq:full_sup_loss}) and compare \model with RandomMask. We also use MOSEI} to assess \modify{how} the number of masked views and the masking ratio \modify{affect} classification performance.

\paragraph{\modify{Loss terms}}
\modify{Table}~\modify{\ref{tab:sup_loss_ablation} compares five loss-term combinations on MOSEI, MAMI, and MUTE.}
\modify{The three-dataset averages are $67.3\%$ with the fused loss only, $66.6\%$ with fused and unimodal losses, $69.6\%$ with fused and masking losses, $69.3\%$ with unimodal and masking losses, and $71.0\%$ with all three terms. Adding the masking term raises the reported average in both comparisons; the full objective has the highest reported mean on each dataset.}
\begin{table}[t]
    \centering
    \caption{
        \modify{Ablation of the supervised objective in Eq.}~\modify{(\ref{eq:full_sup_loss}): classification accuracy (\%). \(\lambda_{\mathrm{fuse}}\), \(\lambda_{\mathrm{uni}}\), and \(\lambda_{\mathrm{mask}}\) indicate whether the fused cross-entropy loss, the auxiliary unimodal cross-entropy losses, and the label-proxy \model loss are included (1) or excluded (0).}
    }
    \label{tab:sup_loss_ablation}
    
    \setlength{\tabcolsep}{4pt}
    \begin{tabular}{ccccccc}
        \toprule
        \modify{\(\lambda_{\mathrm{fuse}}\)} & \modify{\(\lambda_{\mathrm{uni}}\)} & \modify{\(\lambda_{\mathrm{mask}}\)} & \textit{\modify{MOSEI}} & \textit{\modify{MAMI}} & \textit{\modify{MUTE}} & \textit{\modify{Average}} \\
        \midrule
        \modify{1} & \modify{0} & \modify{0} & \modify{$72.8_{\pm0.4}$} & \modify{$64.7_{\pm4.61}$} & \modify{$64.3_{\pm0.97}$} & \modify{$67.3$} \\
        \modify{1} & \modify{1} & \modify{0} & \modify{$72.2_{\pm1.17}$} & \modify{$63.4_{\pm3.44}$} & \modify{$64.1_{\pm2.34}$} & \modify{$66.6$} \\
        \modify{1} & \modify{0} & \modify{1} & \modify{$72.4_{\pm0.2}$} & \modify{$68.7_{\pm1.1}$} & \modify{$67.6_{\pm3.2}$} & \modify{$69.6$} \\
        \modify{0} & \modify{1} & \modify{1} & \modify{$72.4_{\pm0.9}$} & \modify{$66.9_{\pm1.2}$} & \modify{$68.6_{\pm2.0}$} & \modify{$69.3$} \\
        \modify{1} & \modify{1} & \modify{1} & \modify{$73.8_{\pm0.25}$} & \modify{$69.3_{\pm1.60}$} & \modify{$70.0_{\pm2.87}$} & \modify{$71.0$} \\
        \bottomrule
    \end{tabular}
\end{table}

\paragraph{\modify{RandomMask versus \model}}
\modify{Table}~\modify{\ref{tab:sup_masking_ablation} compares Vanilla MT without masking, RandomMask, and \model. RandomMask averages cross-entropy losses over $k$ randomly masked feature views to estimate Eq.}~\modify{(\ref{eq:sup_infmasking}); \model uses the moments of those views in the variational upper bound of Lemma}~\modify{\ref{lemma:sup_ce_gaussian_bound}. RandomMask adds this directly optimized masked-view cross-entropy loss to the unmasked objective.}
\modify{The five-dataset average accuracy increases from $68.1\%$ without masking to $70.8\%$ with RandomMask and $71.7\%$ with \model. \model has higher reported mean accuracy on four datasets and the same rounded mean on MOSEI. The largest differences from RandomMask occur on MAMI and MUTE, at $1.9$ and $1.3$ percentage points, respectively.}
\modify{The supplementary MAMI validation-loss figure plots clean (unmasked) validation cross-entropy for RandomMask and \model in three MAMI runs. Both trajectories fluctuate, while RandomMask reaches higher late-stage peaks in each plotted run.}

\begin{table*}[t]
    \centering
    \caption{
        \modify{RandomMask versus the variational surrogate in the supervised setting (classification accuracy, \%). Average is taken over the five datasets.}
    }
    \label{tab:sup_masking_ablation}
    
    \begin{tabular}{llcccccc}
        \toprule
        & \textit{\modify{Variant}} & \textit{\modify{MOSI}} & \textit{\modify{MOSEI}} & \textit{\modify{MIMIC}} & \textit{\modify{MAMI}} & \textit{\modify{MUTE}} & \textbf{\modify{Average}} \\
        \midrule
        \modify{(a)} & \modify{Vanilla MT (no masking)} & \modify{$72.9_{\pm1.26}$} & \modify{$72.2_{\pm1.17}$} & \modify{$68.1_{\pm0.34}$} & \modify{$63.4_{\pm3.44}$} & \modify{$64.1_{\pm2.34}$} & \modify{$68.1$} \\
        \modify{(b)} & \modify{RandomMask ($k$ masks, no variational bound)} & \modify{$75.3_{\pm0.8}$} & \modify{$\textbf{73.8}_{\pm0.3}$} & \modify{$68.7_{\pm0.5}$} & \modify{$67.4_{\pm2.1}$} & \modify{$68.7_{\pm0.7}$} & \modify{$70.8$} \\
        \modify{(c)} & \modify{\model (variational bound, $k$ masks)} & \modify{$\textbf{75.9}_{\pm0.89}$} & \modify{$\textbf{73.8}_{\pm0.25}$} & \modify{$\textbf{69.4}_{\pm0.06}$} & \modify{$\textbf{69.3}_{\pm1.60}$} & \modify{$\textbf{70.0}_{\pm2.87}$} & \modify{$\textbf{71.7}$} \\
        \bottomrule
    \end{tabular}
\end{table*}

\paragraph{Number of masked views} 
Increasing the number of sampled masked views provides a more accurate finite-sample approximation to the supervised masking objective in Eq.~(\ref{eq:sup_infmasking}), at the expense of additional training cost.
\modify{On MOSEI}, the \modify{supplementary figure shows the highest} accuracy at 7 masked views and a \modify{lower value at 9} views.

\paragraph{Masking ratio} 
The effect of the masking ratio on supervised classification performance is reported in the supplementary material.
\modify{On MOSEI, accuracy} increases \modify{over} the \modify{plotted} masking \modify{ratios} from 40\% to 60\% and then decreases at higher ratios.
This pattern suggests that moderate masking encourages the model to exploit complementary cues across modalities,
whereas overly aggressive masking may remove task-relevant evidence needed for accurate prediction.
Overall,
in the supervised setting,
the optimal masking ratio should balance shortcut suppression against the preservation of discriminative multimodal information.

\paragraph{\modify{Sensitivity to masking hyperparameters}}
\modify{In the supplementary one-factor sweeps, the highest plotted Trifeature synergy accuracy occurs at $k=8$ and $\rho=0.7$; MOSEI accuracy is highest at $k=7$ and $\rho=0.6$. MOSEI accuracy varies by less than one percentage point in each sweep. Each sweep varies one hyperparameter while holding the other fixed.}

\subsection{\modify{Test-time Robustness to a Noisy Dominant Modality}}
\label{sec:robustness}

To further assess the robustness of \model against dominant unimodal noise, we conduct a test-time perturbation experiment on the MOSI dataset. 
We regard the text modality as the dominant unimodal branch, since it achieves the highest accuracy among the three individual modalities. 
Following~\cite{liang2021multibench}, we corrupt the textual input with Gaussian noise at varying levels \modify{to evaluate performance as} the dominant modality \modify{is perturbed}. 

As shown in the supplementary material, the performance of both \model and the Vanilla MT baseline degrades as the noise level increases, but \model retains higher absolute accuracy at every noise level (\textit{e.g.}, 71.7\% vs.\ 66.0\% at level 0.25 and 58.6\% vs.\ 55.8\% at level 1.00).
\modify{In addition to its higher absolute accuracy}, \model shows smaller \modify{relative} performance degradation than Vanilla MT at low-to-moderate noise levels, \modify{while the relative drop becomes comparable} under severe noise (0.75 and 1.00).

\modify{The supplementary robustness table covers MOSI, MOSEI, MAMI, and MUTE, with text and visual corruption at five severities. It includes Gaussian noise, random feature dropout, structured block dropout, and modality dropout. Each perturbation acts on encoded features before fusion, so these results are not directly comparable with the input-level MOSI experiment above. All four datasets use Vanilla MT as the baseline.}

\modify{In the supplementary table, adding \model to Vanilla MT yields higher absolute accuracy in 32 of 32 nonzero-severity entries under Gaussian noise, 27 under modality dropout, 28 under random dropout (with one tie), and 29 under structured dropout. At full modality removal, accuracy is lower for MOSI text (47.33\% versus 53.21\%) and MAMI text (61.53\% versus 64.90\%), and higher for the other six dataset--modality pairs.}

\section{Further Analysis}
\label{sec:additional_exp}

\subsection{\modify{Empirical Tightness of the Variational Bounds}}
\label{sec:tightness_exp}

\modify{For each trained checkpoint, we apply the training-time masking ratio and mask count $k_{\mathrm{train}}$ to each of 100 test inputs. The mask counts range from 4 to 8 across datasets. Using the masked representations produced by the trained model, we compute the surrogate from the empirical moments and directly average the masked objective over the same masks. We report the mean and standard deviation across inputs of the supervised gap $\mathcal{U}-\mathcal{J}$ and the unsupervised gap $\mathcal{J}'-\mathcal{L}'$.}

\modify{The estimated mean gaps are non-negative on all tested datasets at their training-time mask counts. For unsupervised learning, five benchmarks have mean gaps between $0.0006$ and $0.0042$. MUStARD has the largest mean gap and across-input standard deviation, $0.1201\pm0.4755$, at $k_{\mathrm{train}}=5$. The supervised mean gaps are $3.38\times10^{-5}$ on MOSI, $0.0011$ on MOSEI, and $0.0005$ on MIMIC, compared with $0.0645\pm0.0929$ on MAMI and $0.0369\pm0.0926$ on MUTE.}

\modify{Repeating the evaluation with 512 masks per input, with the checkpoints and masking ratios held fixed, yields higher estimated mean gaps on some datasets and lower gaps on others. The MAMI mean gap increases from $0.0645$ at 5 masks to $0.0736$ at 512 masks, whereas the MUStARD mean gap decreases from $0.1201$ at 5 masks to $0.0967$ at 512 masks. At both evaluation counts, MAMI and MUTE have the largest supervised mean gaps, and MUStARD has the largest unsupervised mean gap.}

\modify{In the 512-mask evaluation, the supervised mean gaps on MOSI, MOSEI, and MIMIC are at least two orders of magnitude below their mean masked cross-entropy. On MAMI, the mean gap of $0.0736$ is about $12\%$ of the mean masked cross-entropy, $0.6122$. MAMI and MUTE also have larger covariance scales: $\lambda_{\max}(\Sigma)$ is $1.2449$ and $3.5852$, respectively, compared with $0.0020$ on MOSI. This comparison is consistent with the covariance-dependent analysis in Remark}~\modify{\ref{remark:beyond_gaussian}, which gives two-sided approximation-error bounds for arbitrary mask distributions with finite covariance. The supplementary material provides the complete fixed-checkpoint results, objective and covariance statistics, and sweeps in which each masking ratio and mask count is trained separately.}

\subsection{\modify{Quantifying Synergy in the Learned Representations}}
\label{sec:synergy_quant}

\modify{Trifeature (Section}~\modify{\ref{sec:unsup_exp}) provides constructed synergy labels, and InterSHAP (Section}~\modify{\ref{sec:intershap}) measures interaction attribution in the predictor. For real-world data, we estimate the synergistic information \(S\) of Eq.}~\modify{(\ref{eq:pid}) in learned representations using the information-decomposition method of}~\modify{\cite{liang2023quantifying}. BROJA synergy}~\modify{\cite{bertschinger2014quantifying} is estimated from quantized modality-specific encoder representations and task labels. Values are means with across-seed dispersion in nats; the supplementary material describes the quantization and estimation protocol.}

\begin{table}[t]
    \centering
    \caption{\modify{Estimated synergistic information (in nats) under the unsupervised and supervised settings. Growth rates are computed from unrounded means.}}
    \label{tab:pid_estimates}
    
    \footnotesize
    \setlength{\tabcolsep}{2pt}
    \begin{tabular}{lllcc}
        \toprule
        \textit{\modify{Setting}} & \textit{\modify{Dataset}} & \textit{\modify{Method}} & \textit{\modify{Synergy}} & \makecell{\textit{\modify{Growth}}\\\textit{\modify{(\%)}}} \\
        \midrule
        \multirow{4}{*}{\modify{Unsupervised}}
        & \multirow{2}{*}{\modify{UR-FUNNY}} & \modify{CoMM} & \modify{$0.0362 \pm 0.0037$} & \modify{--} \\
        & & \modify{\model} & \modify{$\mathbf{0.0469} \pm 0.0049$} & \modify{$+29.7\%$} \\
        \cmidrule(lr){2-5}
        & \multirow{2}{*}{\modify{MOSI}} & \modify{CoMM} & \modify{$0.0497 \pm 0.0108$} & \modify{--} \\
        & & \modify{\model} & \modify{$\mathbf{0.0635} \pm 0.0123$} & \modify{$+27.9\%$} \\
        \midrule
        \multirow{4}{*}{\modify{Supervised}}
        & \multirow{2}{*}{\modify{MAMI}} & \modify{Vanilla MT} & \modify{$0.0734 \pm 0.0159$} & \modify{--} \\
        & & \modify{\model} & \modify{$\mathbf{0.0868} \pm 0.0077$} & \modify{$+18.2\%$} \\
        \cmidrule(lr){2-5}
        & \multirow{2}{*}{\modify{MOSEI}} & \modify{Vanilla MT} & \modify{$0.0205 \pm 0.0023$} & \modify{--} \\
        & & \modify{\model} & \modify{$\mathbf{0.0249} \pm 0.0024$} & \modify{$+21.6\%$} \\
        \bottomrule
    \end{tabular}
\end{table}

\modify{Mean estimated synergy under \model is higher than the corresponding baseline on all four benchmarks (Table}~\modify{\ref{tab:pid_estimates}): by $29.7\%$ on UR-FUNNY, $27.9\%$ on MOSI, $18.2\%$ on MAMI, and $21.6\%$ on MOSEI.}

\subsection{\modify{Analysis of Cross-Modal Interactions via InterSHAP}}
\label{sec:intershap}

Beyond downstream task performance, we examine whether \model is associated with a larger contribution from cross-modal interactions to the predictions of the trained models.
We use InterSHAP~\cite{wenderoth2025measuring}, an attribution-based diagnostic that decomposes model behavior into modality-specific main effects and interaction effects using the Shapley Interaction Index\modify{.}
\modify{InterSHAP measures cross-modal interaction attribution in the trained predictor. It differs from the synergistic information \(S\) of Eq.}~\modify{(\ref{eq:pid}), estimated in Section}~\modify{\ref{sec:synergy_quant} from the joint distribution of the quantized modality-specific encoder representations and the task label}.
Let \(\phi_i(x)\) denote the main attribution of modality \(i\) for sample \(x\), and let \(\phi_{ij}(x)\) denote the pairwise interaction attribution between modalities \(i\) and \(j\).
For a test set with \(N\) samples and \(n\) modalities,
the absolute mean interaction between modalities \(i\) and \(j\) is
\begin{equation}
    \Phi_{ij}=\left|\frac{1}{N}\sum_{a=1}^{N}
    \phi_{ij}(m_1^a,\ldots,m_n^a,f)\right|.
    \label{eq:intershap_phi}
\end{equation}
The diagonal term \(\Phi_{ii}\) is defined analogously by replacing \(\phi_{ij}\) with the main attribution \(\phi_i\).
InterSHAP is then the fraction of overall model behavior attributed to cross-modal interactions:
\begin{equation}
    \mathrm{InterSHAP}
    =\frac{\sum_{i\neq j}\Phi_{ij}}{\sum_{i,j}\Phi_{ij}}.
    \label{eq:intershap}
\end{equation}
A larger value indicates that a greater proportion of the model attribution is assigned to joint modality effects rather than to unimodal main effects. 

\begin{table}[t]
    \centering
    \caption{
        Cross-modal interaction analysis with InterSHAP (\%) in the unsupervised (UR-FUNNY, MIMIC) and supervised (MUTE, MAMI) settings.
        Results are reported as the mean \(\pm\) standard deviation over eight independent runs together with paired \(t\)-test results.
    }
    \label{tab:intershap_interaction}
    \setlength{\tabcolsep}{3pt}
    \begin{tabular}{lcccc}
        \toprule
        & \multicolumn{2}{c}{\textit{Unsupervised}} & \multicolumn{2}{c}{\textit{Supervised}} \\
        \cmidrule(lr){2-3}\cmidrule(lr){4-5}
        & \textit{UR-FUNNY} & \textit{MIMIC} & \textit{MUTE} & \textit{MAMI} \\
        \midrule
        w/o \model & $5.30_{\pm1.56}$ & $1.19_{\pm0.66}$ & $0.25_{\pm0.20}$ & $3.81_{\pm2.68}$ \\
        w/ \model & $\mathbf{9.25}_{\pm2.05}$ & $\mathbf{1.67}_{\pm0.68}$ & $\mathbf{8.91}_{\pm6.10}$ & $\mathbf{6.73}_{\pm2.30}$ \\
        \midrule
        Improvement & $+3.95$ & $+0.48$ & $+8.66$ & $+2.92$ \\
        \(t\)/\(p\) & $4.36$ / $0.003$ & $2.61$ / $0.035$ & $3.99$ / $0.005$ & $4.55$ / $0.003$ \\
        \bottomrule
    \end{tabular}
\end{table}

Table~\ref{tab:intershap_interaction} compares models trained without and with \model over eight independent runs with different random seeds.
In the unsupervised setting, \model yields a higher mean interaction attribution on both UR-FUNNY (\(5.30\to9.25\), \(p=0.003\)) and MIMIC (\(1.19\to1.67\), \(p=0.035\)), although the magnitude of the difference varies across datasets.

In the supervised setting, \model likewise raises the mean InterSHAP score on both MUTE (\(0.25\to8.91\), \(p=0.005\)) and MAMI (\(3.81\to6.73\), \(p=0.003\)), \modify{indicating a higher share of} cross-modal \modify{interaction attribution under the evaluated settings}. \modify{Absolute} InterSHAP \modify{levels} remain modest on these real-world benchmarks~\cite{wenderoth2025measuring}.

\modify{On MAMI, every baseline unimodal-sufficiency decile has a positive mean change in per-sample interaction share, though individual changes have both signs. The supplementary material contains the sample-level figure, decile analysis, and estimation protocol.}

\section{\modify{Conclusion}}
\label{sec:conclusion}
\modify{This paper introduces \model, a unified framework for synergistic multimodal learning that stochastically masks modality-specific features before fusion and models an expectation over infinitely many masked views. Contrastive alignment preserves view-level information in the unsupervised setting, whereas label-proxy objectives promote decision consistency in the supervised setting.}
\modify{We further derive tractable variational bounds for the infinite masking objectives, together with tightness guarantees.}
\modify{Across the evaluated benchmarks, \model improves predictive performance in both learning settings. On the controlled Trifeature benchmark, it improves prediction of modality-specific texture and joint texture--color correspondence, supporting the retention of unique and joint evidence. On the analyzed real-world datasets, mean PID estimates indicate higher task-label synergy in learned modality-specific representations, while InterSHAP scores indicate increased predictor interaction attribution.}
\modify{The current analysis justifies the optimization surrogates. Future work will characterize the conditions under which masking recovers synergistic interactions.}
\modify{Training on multiple masked views requires more computation and memory. Performance and resource use with more than three modalities remain untested.}



 

\bibliographystyle{IEEEtran}
\bibliography{main}

@inproceedings{DBLP:conf/iclr/WenZHZX20,
  author       = {Liangjian Wen and
                  Yiji Zhou and
                  Lirong He and
                  Mingyuan Zhou and
                  Zenglin Xu},
  title        = {Mutual Information Gradient Estimation for Representation Learning},
  booktitle    = {Proc. Int. Conf. Learn. Representations},
  year         = {2020}
}

@article{wen2024mveb,
  title={Mveb: Self-supervised learning with multi-view entropy bottleneck},
  author={Wen, Liangjian and Wang, Xiasi and Liu, Jianzhuang and Xu, Zenglin},
  journal={IEEE Trans. Pattern Anal. Mach. Intell.},
  volume={46},
  number={9},
  pages={6097--6108},
  year={2024},
  publisher={IEEE}
}

@inproceedings{wen2025infmasking,
  title={InfMasking: Unleashing Synergistic Information by Contrastive Multimodal Interactions},
  author={Liangjian Wen and Qun Dai and Jianzhuang Liu and Jiangtao Zheng and Yong Dai and Dongkai Wang and Zhao Kang and Jun Wang and Zenglin Xu and Jiang Duan},
  booktitle={Proc. Adv. Neural Inform. Process. Syst.},
  year={2025}
}

@inproceedings{DBLP:conf/iclr/Tsai0SM21,
  author    = {Yao{-}Hung Hubert Tsai and
               Yue Wu and
               Ruslan Salakhutdinov and
               Louis{-}Philippe Morency},
  title     = {Self-supervised Learning from a Multi-view Perspective},
  booktitle = {Proc. Int. Conf. Learn. Representations},
  year      = {2021}}

@inproceedings{DBLP:conf/colt/SridharanK08,
  author    = {Karthik Sridharan and
               Sham M. Kakade},
  title     = {An Information Theoretic Framework for Multi-view Learning},
  booktitle = { Annu. Conf. Learn. Theory},
  year      = {2008}
}

@inproceedings{he2022masked,
  title={Masked autoencoders are scalable vision learners},
  author={He, Kaiming and Chen, Xinlei and Xie, Saining and Li, Yanghao and Doll{\'a}r, Piotr and Girshick, Ross},
  booktitle={Proc. IEEE Conf. Comput. Vis. Pattern Recognit.},
  pages={16000--16009},
  year={2022}
}

@article{bertschinger2014quantifying,
  title={Quantifying unique information},
  author={Bertschinger, Nils and Rauh, Johannes and Olbrich, Eckehard and Jost, J{\"u}rgen and Ay, Nihat},
  journal={Entropy},
  volume={16},
  number={4},
  pages={2161--2183},
  year={2014},
  publisher={Molecular Diversity Preservation International (MDPI)}
}

@article{baltruvsaitis2018multimodal,
  title={Multimodal machine learning: A survey and taxonomy},
  author={Baltru{\v{s}}aitis, Tadas and Ahuja, Chaitanya and Morency, Louis-Philippe},
  journal={IEEE Trans. Pattern Anal. Mach. Intell.},
  volume={41},
  number={2},
  pages={423--443},
  year={2019},
  publisher={IEEE}
}

@inproceedings{hermann2020shapes,
  title={What shapes feature representations? exploring datasets, architectures, and training},
  author={Hermann, Katherine and Lampinen, Andrew},
  booktitle={Proc. Adv. Neural Inform. Process. Syst.},
  volume={33},
  pages={9995--10006},
  year={2020}
}

@inproceedings{liang2021multibench,
  title={{MultiBench}: Multiscale benchmarks for multimodal representation learning},
  author={Liang, Paul Pu and Lyu, Yiwei and Fan, Xiang and Wu, Zetian and Cheng, Yun and Wu, Jason and Chen, Leslie and Wu, Peter and Lee, Michelle A and Zhu, Yuke and others},
  booktitle={Proc. Adv. Neural Inform. Process. Syst. -- Track on Datasets and Benchmarks},
  volume={1},
  year={2021}
}

@inproceedings{arevalo2017gated,
  title={Gated multimodal units for information fusion},
  author={Arevalo, John and Solorio, Thamar and Montes-y-G{\'o}mez, Manuel and Gonz{\'a}lez, Fabio A},
  booktitle={Proc. Int. Conf. Learn. Representations Workshops (ICLR-W)},
  year={2017}
}

@article{johnson2016mimic,
  title={{MIMIC-III, a freely accessible critical care database}},
  author={Johnson, Alistair EW and Pollard, Tom J and Shen, Lu and Lehman, Li-wei H and Feng, Mengling and Ghassemi, Mohammad and Moody, Benjamin and Szolovits, Peter and Anthony Celi, Leo and Mark, Roger G},
  journal={Scientific data},
  volume={3},
  number={1},
  pages={1--9},
  year={2016},
  publisher={Nature Publishing Group}
}

@article{zadeh2016multimodal,
  title={Multimodal sentiment intensity analysis in videos: Facial gestures and verbal messages},
  author={Zadeh, Amir and Zellers, Rowan and Pincus, Eli and Morency, Louis-Philippe},
  journal={IEEE Intell. Syst.},
  volume={31},
  number={6},
  pages={82--88},
  year={2016},
  publisher={IEEE}
}

@inproceedings{radford2021learning,
  title={Learning transferable visual models from natural language supervision},
  author={Radford, Alec and Kim, Jong Wook and Hallacy, Chris and Ramesh, Aditya and Goh, Gabriel and Agarwal, Sandhini and Sastry, Girish and Askell, Amanda and Mishkin, Pamela and Clark, Jack and Krueger, Gretchen and Sutskever, Ilya },
  booktitle={Proc. Int. Conf. Mach. Learn.},
  pages={8748--8763},
  year={2021},
}

@inproceedings{jia2021scaling,
  title={Scaling up visual and vision-language representation learning with noisy text supervision},
  author={Jia, Chao and Yang, Yinfei and Xia, Ye and Chen, Yi-Ting and Parekh, Zarana and Pham, Hieu and Le, Quoc and Sung, Yun-Hsuan and Li, Zhen and Duerig, Tom},
  booktitle={Proc. Int. Conf. Mach. Learn.},
  pages={4904--4916},
  year={2021},
}

@inproceedings{mu2022slip,
  title={{SLIP: Self-supervision meets language-image pre-training}},
  author={Mu, Norman and Kirillov, Alexander and Wagner, David and Xie, Saining},
  booktitle={Proc. Eur. Conf. Comput. Vis.},
  pages={529--544},
  year={2022}
}

@inproceedings{liang2023quantifying,
  title={Quantifying \& modeling multimodal interactions: An information decomposition framework},
  author={Liang, Paul Pu and Cheng, Yun and Fan, Xiang and Ling, Chun Kai and Nie, Suzanne and Chen, Richard and Deng, Zihao and Allen, Nicholas and Auerbach, Randy and Mahmood, Faisal and Salakhutdinov, Ruslan and Morency, Louis-Philippe},
  booktitle={Proc. Adv. Neural Inform. Process. Syst.},
  volume={36},
  pages = {27351--27393},
  year={2023}
}

@inproceedings{tian2020cmc,
  title={Contrastive Multiview Coding},
  author={Tian, Yonglong and Krishnan, Dilip and Isola, Phillip},
  booktitle={Proc. Eur. Conf. Comput. Vis.},
  pages={776--794},
  year={2020}
}

@inproceedings{federici2020learning,
  title={Learning robust representations via multi-view information bottleneck},
  author={Federici, Marco and Dutta, Anjan and Forr{\'e}, Patrick and Kushman, Nate and Akata, Zeynep},
  booktitle={Proc. Int. Conf. Learn. Representations},
  year={2020}
}

@inproceedings{liang2023factorized,
  title={Factorized Contrastive Learning: Going Beyond Multi-view Redundancy},
  author={Liang, Paul Pu and Deng, Zihao and Ma, Martin and Zou, James and Morency, Louis-Philippe and Salakhutdinov, Ruslan},
  booktitle={Proc. Adv. Neural Inform. Process. Syst.},
  volume={36},
  pages = {32971--32998},
  year={2023}
}

@inproceedings{bachmann2022multimae,
  author    = {Roman Bachmann and David Mizrahi and Andrei Atanov and Amir Zamir},
  title     = {{MultiMAE}: Multi-modal Multi-task Masked Autoencoders},
  booktitle   = {Proc. Eur. Conf. Comput. Vis.},
  pages =   {348--367},
  year      = {2022},
}

@inproceedings{yuan2021multimodal,
  title={Multimodal contrastive training for visual representation learning},
  author={Yuan, Xin and Lin, Zhe and Kuen, Jason and Zhang, Jianming and Wang, Yilin and Maire, Michael and Kale, Ajinkya and Faieta, Baldo},
  booktitle={Proc. IEEE Conf. Comput. Vis. Pattern Recognit.},
  pages={6995--7004},
  year={2021}
}

@inproceedings{chen2020simple,
  title={A simple framework for contrastive learning of visual representations},
  author={Chen, Ting and Kornblith, Simon and Norouzi, Mohammad and Hinton, Geoffrey},
  booktitle={Proc. Int. Conf. Mach. Learn.},
  pages={1597--1607},
  year={2020}
}

@article{xu2023multimodal,
  title={Multimodal learning with transformers: A survey},
  author={Xu, Peng and Zhu, Xiatian and Clifton, David A},
  journal={IEEE Trans. Pattern Anal. Mach. Intell.},
  volume={45},
  number={10},
  pages={12113--12132},
  year={2023},
  publisher={IEEE}
}

@article{williams2010nonnegative,
  title={Nonnegative decomposition of multivariate information},
  author={Williams, Paul L and Beer, Randall D},
  journal={arXiv preprint arXiv:1004.2515},
  year={2010}
}

@article{barnum2020benefits,
  title={On the benefits of early fusion in multimodal representation learning},
  author={Barnum, George and Talukder, Sabera and Yue, Yisong},
  journal={arXiv preprint arXiv:2011.07191},
  year={2020}
}

@inproceedings{tsai2019multimodal,
  title={Multimodal transformer for unaligned multimodal language sequences},
  author={Tsai, Yao-Hung Hubert and Bai, Shaojie and Liang, Paul Pu and Kolter, J Zico and Morency, Louis-Philippe and Salakhutdinov, Ruslan},
  booktitle={Proc. 57th Annu. Meeting Assoc. Comput. Linguistics},
  pages={6558--6569},
  year={2019}
}

@inproceedings{dufumier2024align,
  title={What to align in multimodal contrastive learning?},
  author={Dufumier, Benoit and Castillo-Navarro, Javiera and Tuia, Devis and Thiran, Jean-Philippe},
  booktitle={Proc. Int. Conf. Learn. Representations},
  year={2025}
}

@inproceedings{wenderoth2025measuring,
  title={Measuring cross-modal interactions in multimodal models},
  author={Wenderoth, Laura and Hemker, Konstantin and Simidjievski, Nikola and Jamnik, Mateja},
  booktitle={Proc. AAAI Conf. Artif. Intell.},
  volume={39},
  pages={21501--21509},
  year={2025}
}

@article{lin2022dual,
  title={Dual contrastive prediction for incomplete multi-view representation learning},
  author={Lin, Yijie and Gou, Yuanbiao and Liu, Xiaotian and Bai, Jinfeng and Lv, Jiancheng and Peng, Xi},
  journal={IEEE Trans. Pattern Anal. Mach. Intell.},
  volume={45},
  number={4},
  pages={4447--4461},
  year={2023},
  publisher={IEEE}
}

@article{guo2019deep,
  title={Deep multimodal representation learning: A survey},
  author={Guo, Wenzhong and Wang, Jianwen and Wang, Shiping},
  journal={IEEE Access},
  volume={7},
  pages={63373--63394},
  year={2019},
  publisher={IEEE}
}

@inproceedings{nagrani2021attention,
  title={Attention bottlenecks for multimodal fusion},
  author={Nagrani, Arsha and Yang, Shan and Arnab, Anurag and Jansen, Aren and Schmid, Cordelia and Sun, Chen},
  booktitle={Proc. Adv. Neural Inform. Process. Syst.},
  volume={34},
  pages={14200--14213},
  year={2021}
}

@inproceedings{wu2024multimodal,
  title={Multimodal multi-loss fusion network for sentiment analysis},
  author={Wu, Zehui and Gong, Ziwei and Koo, Jaywon and Hirschberg, Julia},
  booktitle={Proc. Conf. North Amer. Chapter Assoc. Comput. Linguist.},
  pages={3588--3602},
  year={2024}
}

@inproceedings{lu2019vilbert,
  title={Vilbert: Pretraining task-agnostic visiolinguistic representations for vision-and-language tasks},
  author={Lu, Jiasen and Batra, Dhruv and Parikh, Devi and Lee, Stefan},
  booktitle={Proc. Adv. Neural Inform. Process. Syst.},
  volume={32},
  year={2019}
}

@inproceedings{sun2019videobert,
  title={Videobert: A joint model for video and language representation learning},
  author={Sun, Chen and Myers, Austin and Vondrick, Carl and Murphy, Kevin and Schmid, Cordelia},
  booktitle={Proc. IEEE Int. Conf. Comput. Vis.},
  pages={7464--7473},
  year={2019}
}

@inproceedings{chen2020uniter,
  title={Uniter: Universal image-text representation learning},
  author={Chen, Yen-Chun and Li, Linjie and Yu, Licheng and El Kholy, Ahmed and Ahmed, Faisal and Gan, Zhe and Cheng, Yu and Liu, Jingjing},
  booktitle={Proc. Eur. Conf. Comput. Vis.},
  pages={104--120},
  year={2020},
}

@inproceedings{wang2020makes,
  title={What makes training multi-modal classification networks hard?},
  author={Wang, Weiyao and Tran, Du and Feiszli, Matt},
  booktitle={Proc. IEEE Conf. Comput. Vis. Pattern Recognit.},
  pages={12695--12705},
  year={2020}
}

@inproceedings{yao2022modality,
  title={Modality-specific learning rates for effective multimodal additive late-fusion},
  author={Yao, Yiqun and Mihalcea, Rada},
  booktitle={Findings Assoc. Comput. Linguist.},
  pages={1824--1834},
  year={2022}
}

@inproceedings{li2020oscar,
  title={Oscar: Object-semantics aligned pre-training for vision-language tasks},
  author={Li, Xiujun and Yin, Xi and Li, Chunyuan and Zhang, Pengchuan and Hu, Xiaowei and Zhang, Lei and Wang, Lijuan and Hu, Houdong and Dong, Li and Wei, Furu and others},
  booktitle={Proc. Eur. Conf. Comput. Vis.},
  pages={121--137},
  year={2020},
}

@inproceedings{guo2025disentangling,
  title={Disentangling homophily and heterophily in multimodal graph clustering},
  author={Guo, Zhaochen and Shen, Zhixiang and Xie, Xuanting and Wen, Liangjian and Kang, Zhao},
  booktitle={Proc. 33rd ACM Int. Conf. Multimedia},
  pages={2044--2053},
  year={2025}
}

@inproceedings{wang2024decoupling,
  title={Decoupling common and unique representations for multimodal self-supervised learning},
  author={Wang, Yi and Albrecht, Conrad M and Braham, Nassim Ait Ali and Liu, Chenying and Xiong, Zhitong and Zhu, Xiao Xiang},
  booktitle={Proc. Eur. Conf. Comput. Vis.},
  pages={286--303},
  year={2024},
}

@inproceedings{peng2022balanced,
  title={Balanced multimodal learning via on-the-fly gradient modulation},
  author={Peng, Xiaokang and Wei, Yake and Deng, Andong and Wang, Dong and Hu, Di},
  booktitle={Proc. IEEE Conf. Comput. Vis. Pattern Recognit.},
  pages={8238--8247},
  year={2022}
}

@inproceedings{zhou2023intra,
  title={Intra-and inter-modal curriculum for multimodal learning},
  author={Zhou, Yuwei and Wang, Xin and Chen, Hong and Duan, Xuguang and Zhu, Wenwu},
  booktitle={Proc. 31st ACM Int. Conf. Multimedia},
  pages={3724--3735},
  year={2023}
}

@article{stahlschmidt2022multimodal,
  title={Multimodal deep learning for biomedical data fusion: a review},
  author={Stahlschmidt, S{\"o}ren Richard and Ulfenborg, Benjamin and Synnergren, Jane},
  journal={Briefings in bioinformatics},
  volume={23},
  number={2},
  pages={bbab569},
  year={2022},
  publisher={Oxford University Press}
}

@article{liang2024foundations,
  title={Foundations \& trends in multimodal machine learning: Principles, challenges, and open questions},
  author={Liang, Paul Pu and Zadeh, Amir and Morency, Louis-Philippe},
  journal={ACM Comput. Surv.},
  volume={56},
  number={10},
  pages={1--42},
  year={2024},
  publisher={ACM New York, NY}
}

@inproceedings{zhang2024multimodal,
  title={Multimodal representation learning by alternating unimodal adaptation},
  author={Zhang, Xiaohui and Yoon, Jaehong and Bansal, Mohit and Yao, Huaxiu},
  booktitle={Proc. IEEE Conf. Comput. Vis. Pattern Recognit.},
  pages={27456--27466},
  year={2024}
}

@inproceedings{chaudhuri2025closer,
  title={A closer look at multimodal representation collapse},
  author={Chaudhuri, Abhra and Dutta, Anjan and Bui, Tu and Georgescu, Serban},
  booktitle={Proc. Int. Conf. Mach. Learn.},
  year={2025}
}

@article{oord2018representation,
  title={Representation learning with contrastive predictive coding},
  author={Oord, Aaron van den and Li, Yazhe and Vinyals, Oriol},
  journal={arXiv preprint arXiv:1807.03748},
  year={2018}
}

@inproceedings{huang2022modality,
  title={Modality competition: What makes joint training of multi-modal network fail in deep learning? (provably)},
  author={Huang, Yu and Lin, Junyang and Zhou, Chang and Yang, Hongxia and Huang, Longbo},
  booktitle={Proc. Int. Conf. Mach. Learn.},
  pages={9226--9259},
  year={2022},
}

@inproceedings{vielzeuf2018centralnet,
  title={Centralnet: a multilayer approach for multimodal fusion},
  author={Vielzeuf, Valentin and Lechervy, Alexis and Pateux, St{\'e}phane and Jurie, Fr{\'e}d{\'e}ric},
  booktitle={Proc. Eur. Conf. Comput. Vis. Workshops},
  year={2018}
}

@inproceedings{xu2023mmcosine,
  title={Mmcosine: Multi-modal cosine loss towards balanced audio-visual fine-grained learning},
  author={Xu, Ruize and Feng, Ruoxuan and Zhang, Shi-Xiong and Hu, Di},
  booktitle={Proc. IEEE Int. Conf. Acoust. Speech Signal Process.},
  pages={1--5},
  year={2023},
}

@article{wei2024mmpareto,
  title={Mmpareto: Boosting multimodal learning with innocent unimodal assistance},
  author={Wei, Yake and Hu, Di},
  journal={arXiv preprint arXiv:2405.17730},
  year={2024}
}

@inproceedings{wei2024diagnosing,
  title={Diagnosing and re-learning for balanced multimodal learning},
  author={Wei, Yake and Li, Siwei and Feng, Ruoxuan and Hu, Di},
  booktitle={Proc. Eur. Conf. Comput. Vis.},
  pages={71--86},
  year={2024},
}

@inproceedings{cai2020joint,
  title={Joint contrastive learning with infinite possibilities},
  author={Cai, Qi and Wang, Yu and Pan, Yingwei and Yao, Ting and Mei, Tao},
  booktitle={Proc. Adv. Neural Inform. Process. Syst.},
  volume={33},
  pages={12638--12648},
  year={2020}
}

@inproceedings{waligora2024joint,
  title={Joint multimodal transformer for emotion recognition in the wild},
  author={Waligora, Paul and Aslam, Muhammad Haseeb and Zeeshan, Muhammad Osama and Belharbi, Soufiane and Koerich, Alessandro Lameiras and Pedersoli, Marco and Bacon, Simon and Granger, Eric},
  booktitle={Proc. IEEE Conf. Comput. Vis. Pattern Recognit. Workshops},
  pages={4625--4635},
  year={2024}
}

@inproceedings{fersini2022semeval,
  title={SemEval-2022 task 5: Multimedia automatic misogyny identification},
  author={Fersini, Elisabetta and Gasparini, Francesca and Rizzi, Giulia and Saibene, Aurora and Chulvi, Berta and Rosso, Paolo and Lees, Alyssa and Sorensen, Jeffrey},
  booktitle={Proc. 16th Int. Workshop Semantic Eval.},
  pages={533--549},
  year={2022}
}

@inproceedings{hossain2022mute,
  title={MUTE: A multimodal dataset for detecting hateful memes},
  author={Hossain, Eftekhar and Sharif, Omar and Hoque, Mohammed Moshiul},
  booktitle={Proc. 2nd Conf. Asia-Pacific Chapter Assoc. Comput. Linguist.: Student Res. Workshop},
  pages={32--39},
  year={2022}
}

@inproceedings{zadeh2018multimodal,
  title={Multimodal language analysis in the wild: Cmu-mosei dataset and interpretable dynamic fusion graph},
  author={Zadeh, AmirAli Bagher and Liang, Paul Pu and Poria, Soujanya and Cambria, Erik and Morency, Louis-Philippe},
  booktitle={Proc. 56th Annu. Meeting Assoc. Comput. Linguist.},
  pages={2236--2246},
  year={2018}
}

@article{guo2026sama,
  title={{SAMA}: Semantic Anchor-Aligned Augmentation for Unified Low-Resource Multimodal Information Extraction},
  author={Guo, Quanjiang and Mu, Chong and Pan, Jiazhou and Jia, Ming and Tian, Ling and Gao, Hui and Kang, Zhao},
  journal={IEEE Trans. Multimedia},
  volume={28},
  pages={13167--13179},
  year={2026}
}

@inproceedings{wang2025cooperation,
  title={Cooperation of Experts: Fusing Heterogeneous Information with Large Margin},
  author={Wang, Shuo and Huang, Shunyang and Yuan, Jinghui and Shen, Zhixiang and Kang, Zhao},
  booktitle={Proc. Int. Conf. Mach. Learn.},
  series={Proceedings of Machine Learning Research},
  volume={267},
  pages={63169--63185},
  year={2025}
}

@inproceedings{dai2026hril,
  title={{HRIL}: Learning Multimodal Synergy via Higher-Order Tensor Modeling},
  author={Dai, Qun and Wen, Liangjian and Duan, Jiang and Dai, Yong and Wang, Dongkai and Wang, Maolin and Wang, Mingjie and Liu, Jianzhuang and Yan, He and Kang, Zhao},
  booktitle={Proc. Adv. Neural Inform. Process. Syst.},
  year={2026}
}

@inproceedings{chen2026mscr,
  title={{MSCR}: Jointly Balancing Utilization Across Modalities and Discovering Synergistic Information},
  author={Chen, Xinyu and Wen, Liangjian and Duan, Jiang and Wang, Dongkai and Dai, Yong and Bai, Jiayu and Liu, Jianzhuang and Tian, Zhen and Qiu, Guoping and Kang, Zhao},
  booktitle={Proc. Adv. Neural Inform. Process. Syst.},
  year={2026}
}

@inproceedings{guo2026webcogreasoner,
  title={{Web-CogReasoner}: Towards Multimodal Knowledge-Induced Cognitive Reasoning for Web Agents},
  author={Guo, Yuhan and Guo, Cong and Sun, Aiwen and He, Hongliang and Yang, Xinyu and Lu, Yue and Zhang, Yingji and Guo, Xuntao and Zhang, Dong and Liu, Jianzhuang and Duan, Jiang and Xiao, Yijia and Wen, Liangjian and Xu, Hai-Ming and Dai, Yong},
  booktitle={Proc. Int. Conf. Learn. Representations},
  pages={36186--36210},
  year={2026}
}

@article{wen2026dependency,
  title={Dependency, Compression, and Synergy: A Unified Information-Theoretic View of Multimodal Learning},
  author={Wen, Liangjian and Li, Linjie and Duan, Jiang and Dai, Yong and Liu, Jianzhuang and Kang, Zhao},
  journal={arXiv preprint arXiv:2609.14421},
  year={2026}
}

@article{lou2026granular,
  title={Granular Information Bottleneck for Deep Multi-Modal Clustering},
  author={Lou, Zhengzheng and Lv, Mingyang and Zhan, Yuhan and Li, Yingxuan and Zhang, Chaoyang and Hu, Shizhe},
  journal={IEEE Trans. Image Process.},
  volume={35},
  pages={7307--7318},
  year={2026},
  publisher={IEEE}
}

@article{hu2026trust,
  title={To the Best of Trust: Full-Stage Trusted Multi-Modal Clustering},
  author={Hu, Shizhe and Fan, Jiahao and Wu, Yucong and Wang, Jinlan and Jiang, Xiaoheng and Lv, Pei and Xu, Mingliang},
  journal={IEEE Trans. Image Process.},
  volume={35},
  pages={8635--8647},
  year={2026},
  publisher={IEEE}
}

@article{hu2025pseudolabel,
  title={Deep Multiview Clustering by Pseudo-Label Guided Contrastive Learning and Dual Correlation Learning},
  author={Hu, Shizhe and Zhang, Chengkun and Zou, Guoliang and Lou, Zhengzheng and Ye, Yangdong},
  journal={IEEE Trans. Neural Netw. Learn. Syst.},
  volume={36},
  number={2},
  pages={3646--3658},
  year={2025},
  publisher={IEEE}
}

@article{lou2026structure,
  title={Structure-Enhanced Self-Supervised Weighted Information Bottleneck for Multiview Clustering},
  author={Lou, Zhengzheng and Wu, Yucong and Zhang, Ke and Zhang, Chaoyang and Ji, Bo and Hu, Shizhe},
  journal={IEEE Trans. Neural Netw. Learn. Syst.},
  volume={37},
  number={8},
  pages={3727--3740},
  year={2026},
  publisher={IEEE}
}

@article{hu2024survey,
  title={A Survey on Information Bottleneck},
  author={Hu, Shizhe and Lou, Zhengzheng and Yan, Xiaoqiang and Ye, Yangdong},
  journal={IEEE Trans. Pattern Anal. Mach. Intell.},
  volume={46},
  number={8},
  pages={5325--5344},
  year={2024},
  publisher={IEEE}
}

@article{hu2024multiview,
  title={Multiview clustering with propagating information bottleneck},
  author={Hu, Shizhe and Shi, Zenglin and Yan, Xiaoqiang and Lou, Zhengzheng and Ye, Yangdong},
  journal={IEEE Trans. Neural Netw. Learn. Syst.},
  volume={35},
  number={7},
  pages={9915--9929},
  year={2024},
  publisher={IEEE}
}

@article{zhang2023multi,
  title={Multi-view clustering via triplex information maximization},
  author={Zhang, Chaoyang and Lou, Zhengzheng and Zhou, Qinglei and Hu, Shizhe},
  journal={IEEE Trans. Image Process.},
  volume={32},
  pages={4299--4313},
  year={2023},
  publisher={IEEE}
}

@inproceedings{kiela2020hateful,
  title={The hateful memes challenge: Detecting hate speech in multimodal memes},
  author={Kiela, Douwe and Firooz, Hamed and Mohan, Aravind and Goswami, Vedanuj and Singh, Amanpreet and Ringshia, Pratik and Testuggine, Davide},
  booktitle={Proc. Adv. Neural Inform. Process. Syst.},
  volume={33},
  pages={2611--2624},
  year={2020}
}

\makeatletter
\def\@textbottom{\vskip \z@ \@plus 1fill}
\let\@texttop\relax
\def\@IEEEBIOskipN{0.3\baselineskip}
\makeatother
\makeatletter
\def\@IEEEBIOphotowidth{0.85in}
\def\@IEEEBIOphotodepth{1.0625in}
\def\@IEEEBIOhangwidth{0.99in}
\def\@IEEEBIOhangdepth{1.0625in}
\makeatother

\begin{IEEEbiography}[{\includegraphics[width=0.85in,height=1.0625in,clip,keepaspectratio]{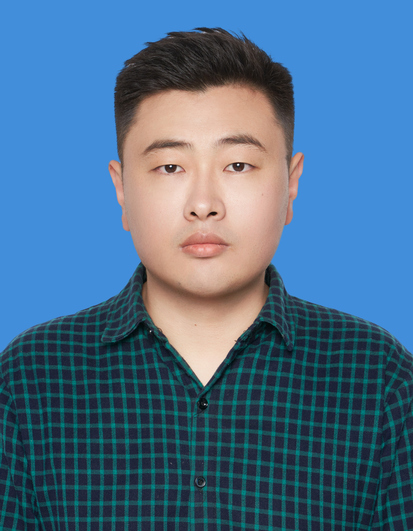}}]{Liangjian Wen} is currently an associate professor with the School of Computing and Artificial Intelligence, Southwestern University of Finance and Economics, Chengdu, China. He received his Ph.D. degree from the University of Electronic Science and Technology of China in July 2021. His research interests include machine learning, deep learning, representation learning, and self-supervised learning.
\end{IEEEbiography}

\begin{IEEEbiography}[{\includegraphics[width=0.85in,height=1.0625in,clip,keepaspectratio]{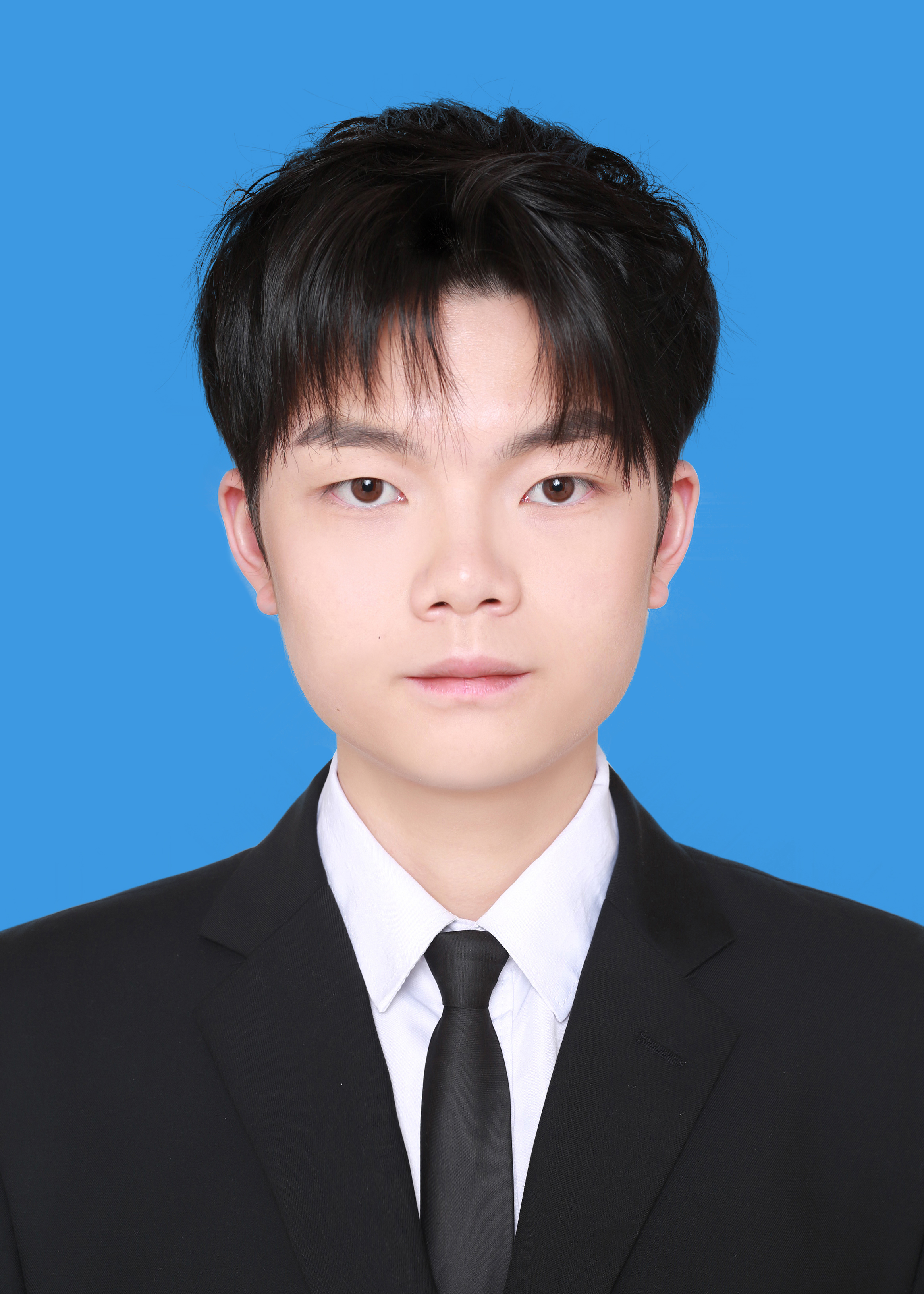}}]{Qun Dai} received the B.E. degree from Southwestern University of Finance and Economics, Chengdu, China, in 2025. He is currently working toward the master's degree with the School of Computing and Artificial Intelligence, Southwestern University of Finance and Economics, Chengdu, China. His research interests include machine learning, deep learning, and representation learning.
\end{IEEEbiography}

\begin{IEEEbiography}[{\includegraphics[width=0.85in,height=1.0625in, clip,keepaspectratio]{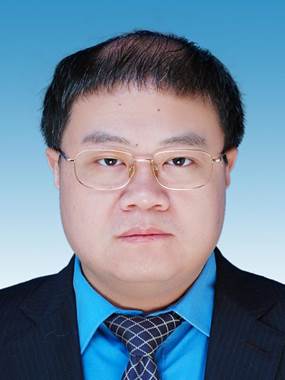}}]{Jiang Duan} received the BSc degree in mechanical engineering from Southwest Jiaotong University, Chengdu, China, in 2001, the MSc degree from the University of Derby, Derby, U.K., in 2002, and the Ph.D. degree from the University of Nottingham, U.K., in 2006. He is a professor at the Southwestern University of Finance and Economics, China. He is a recognized national expert, the winner of the first Sichuan Outstanding Talent Award (The Highest Talent Award of Sichuan Province), the winner of the Sichuan Youth Science and Technology Award, and a standing committee member of the Sichuan Association for Science and Technology.
\end{IEEEbiography}

\begin{IEEEbiography}[{\includegraphics[width=0.85in,height=1.0625in,clip,keepaspectratio]{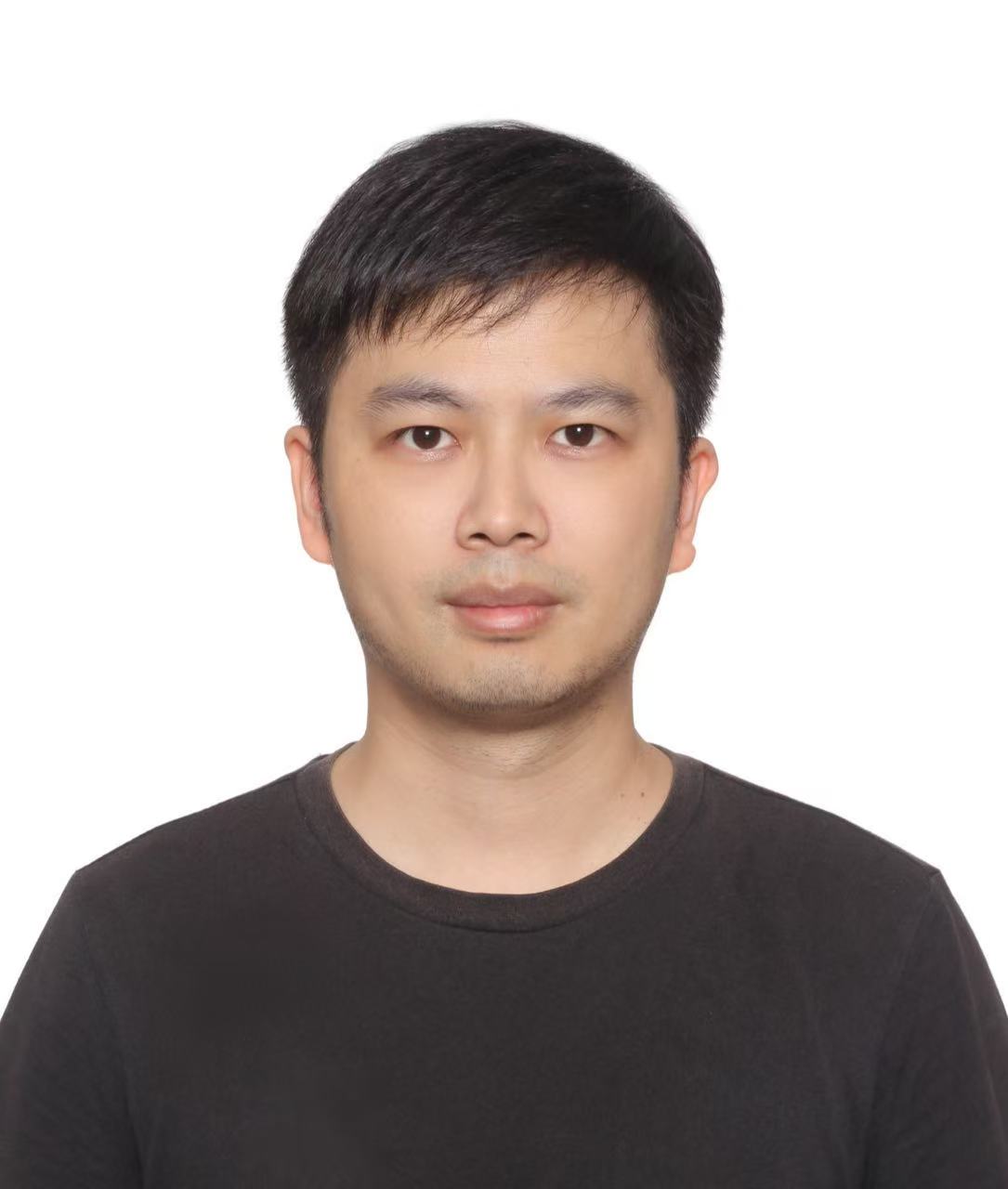}}]{Yong Dai} received his Ph.D. from the University of Electronic Science and Technology of China in 2021. He previously conducted research on large language models at Microsoft Research Asia and at Tencent AI Lab. Since May 2025, he has served as Head of Multimodal Large Models at the Beijing Humanoid Robot Innovation Center, leading research on embodied brain-cerebellum systems and world models.
\end{IEEEbiography}

\begin{IEEEbiography}[{\includegraphics[width=0.85in,height=1.0625in,clip,keepaspectratio]{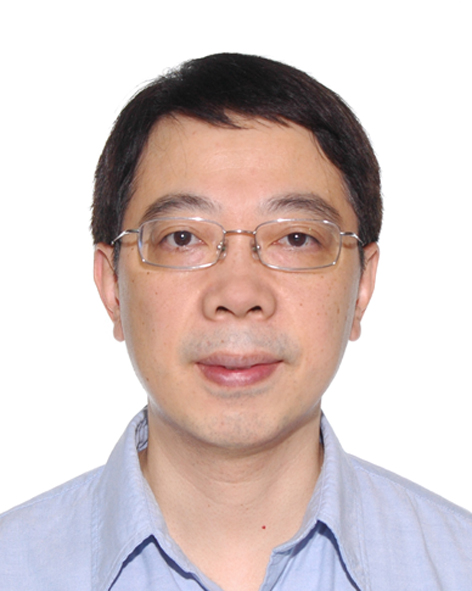}}]{Jianzhuang Liu} (Senior Member, IEEE) received the PhD degree in computer vision from The Chinese University of Hong Kong, in 1997. From 1998 to 2000, he was a research fellow with Nanyang Technological University, Singapore. From 2000 to 2012, he was a postdoctoral fellow, an assistant professor, and an adjunct associate professor with The Chinese University of Hong Kong, Hong Kong. In 2011, he joined the Shenzhen Institutes of Advanced Technology, Chinese Academy of Sciences, Shenzhen, China, as a professor. He was a principal researcher in Huawei Company from 2012 to 2023. He has authored more than 200 papers in the areas of computer vision, image processing, machine learning, and multimodal AI.
\end{IEEEbiography}

\begin{IEEEbiography}[{\includegraphics[width=0.85in,height=1.0625in,clip,keepaspectratio]{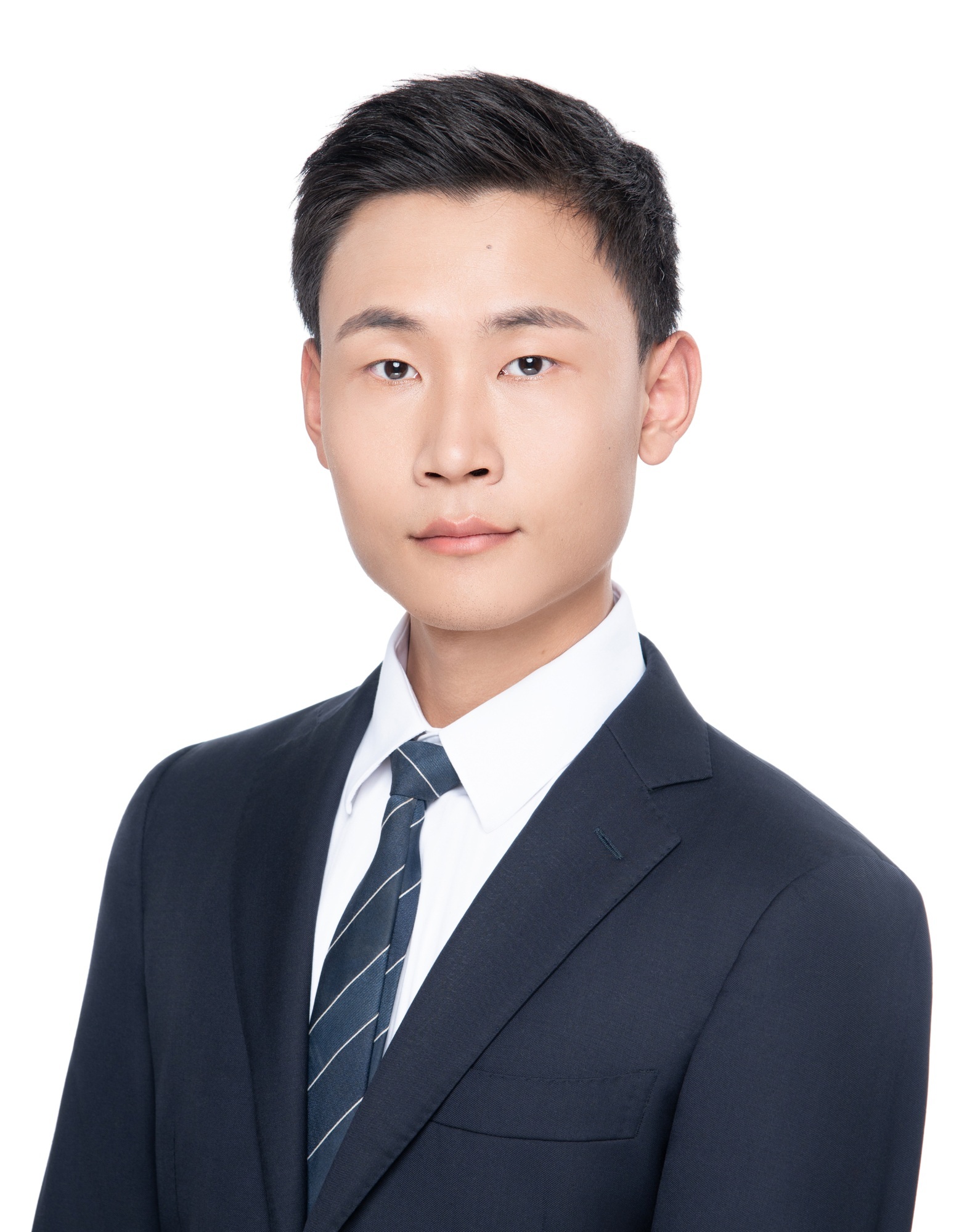}}]
{Dongkai Wang} received the Ph.D. degree from Peking University and the B.E. degree from the University of Electronic Science and Technology of China. He is currently a professor at Southwestern University of Finance and Economics. His research focuses on human-aware AI, including fine-grained perception, human motion reconstruction, and motion generation. He has received doctoral dissertation awards from CSIG and Peking University and serves on the CCF/CSIG Technical Committee on Multimedia.
\end{IEEEbiography}

\begin{IEEEbiography}[{\includegraphics[width=0.85in,height=1.0625in,clip,keepaspectratio]{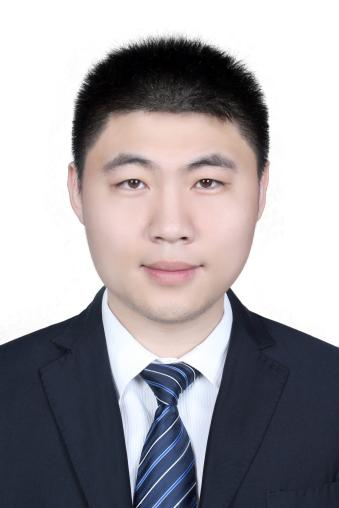}}]{Jun Wang} is a Professor and Doctoral Supervisor at the School of Management Science and Engineering, Southwestern University of Finance and Economics. Jun Wang has long \modify{conducted} interdisciplinary research in fintech, financial intelligence, and digital-intelligence management, using machine learning, data analytics, and artificial intelligence methods \modify{to study financial markets from a big data analytics perspective}. 
\end{IEEEbiography}

\begin{IEEEbiography}[{\includegraphics[width=0.85in,height=1.0625in,clip,keepaspectratio]{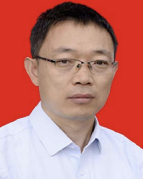}}]{Fuyuan Cao} received the M.S. and Ph.D. degrees in computer science from Shanxi University, Taiyuan, China, in 2004 and 2010, respectively. He is currently a professor with the School of Computer and Information Technology, Shanxi University, China. His current research interests include machine learning and clustering analysis.
\end{IEEEbiography}

\begin{IEEEbiography}[{\includegraphics[width=0.85in,height=1.0625in,clip,keepaspectratio]{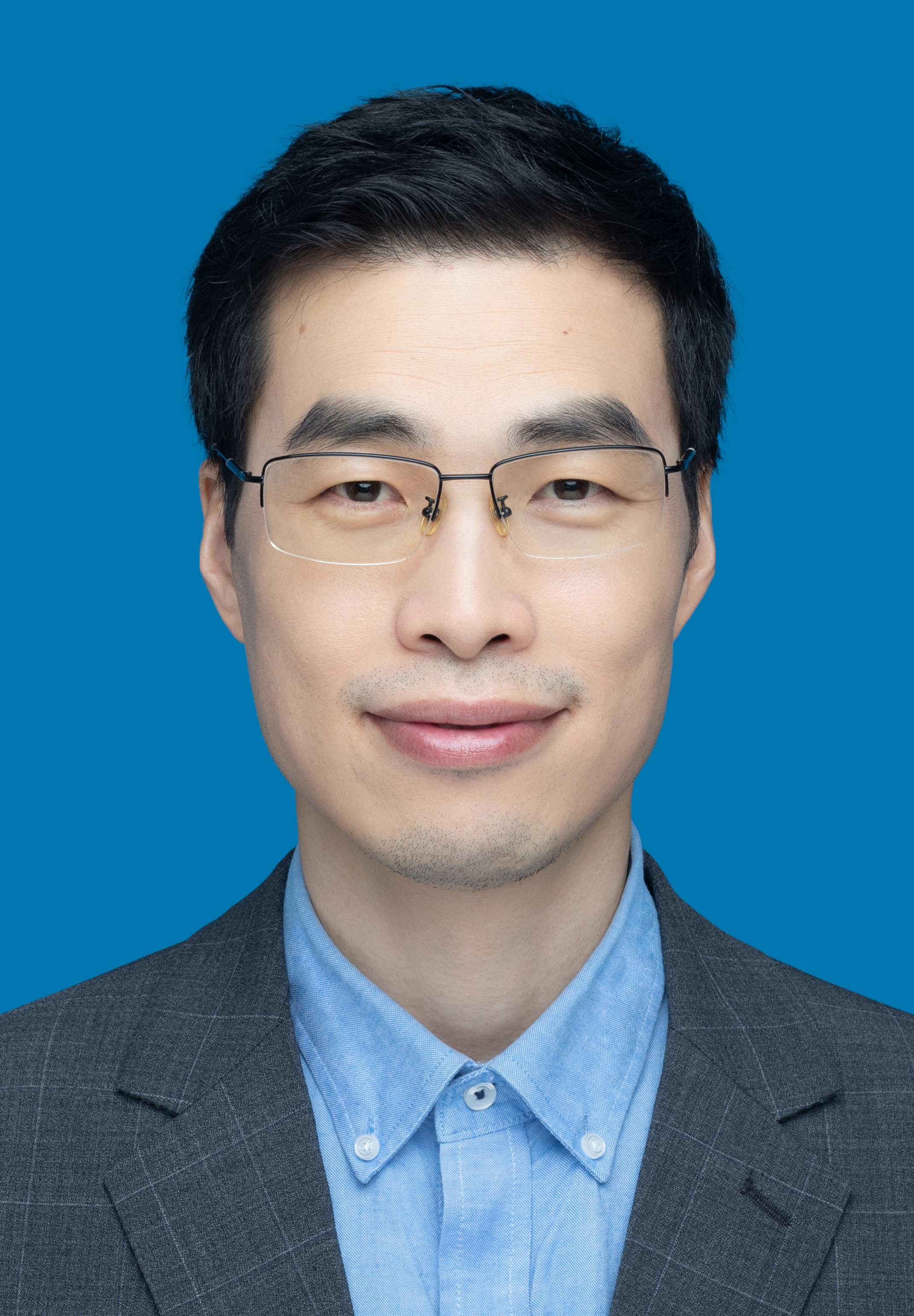}}]{Zenglin Xu} (SM'20) is a Distinguished Professor at the Artificial Intelligence Innovation and Incubation (AI\textsuperscript{3}) Institute, Fudan University, and an AI Scientist at the Shanghai Academy of AI for Science. He earned his Ph.D. in Computer Science and Engineering from The Chinese University of Hong Kong and completed postdoctoral research at Purdue University and the Max Planck Institute for Informatics. Dr. Xu serves as Vice President of the International Neural Network Society (INNS) and is the Editor-in-Chief of Academia AI and Applications. His research centers on trustworthy machine learning and agentic AI, AI for Science, and domain-specific applications in healthcare, finance and education.
\end{IEEEbiography}

\begin{IEEEbiography}[{\includegraphics[width=0.85in,height=1.0625in,clip,keepaspectratio]{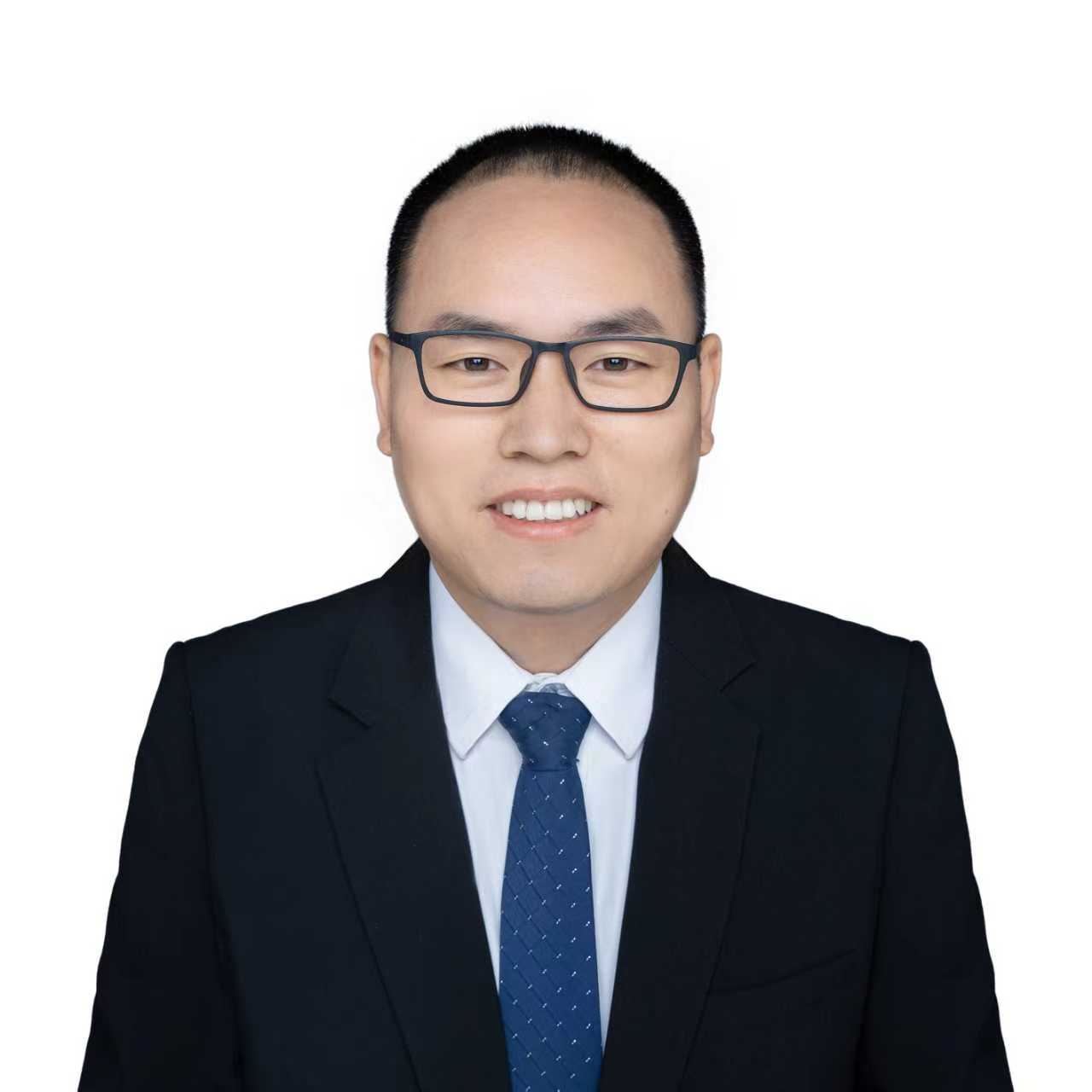}}]{Zhao Kang} received the Ph.D. degree in computer science from Southern Illinois University Carbondale, Carbondale, IL, USA, in 2017. He is currently a Professor in the School of Computer Science and Engineering, University of Electronic Science and Technology of China, Chengdu, China. His research interests include graph machine learning and large language models. He serves as an Associate Editor for Neural Networks and Pattern Recognition.
\end{IEEEbiography}

\ifdefined\INCLUDEAPPENDIX
\clearpage
\flushbottom
\input{section/appendix.tex}
\fi

\end{document}